\documentclass[conference]{IEEEtran}
\IEEEoverridecommandlockouts
\usepackage{cite}
\usepackage{amsmath,amssymb,amsfonts}
\usepackage{algorithmic}
\usepackage[ruled,linesnumbered,noline,noend]{algorithm2e}
\usepackage{threeparttable}
\usepackage{booktabs}
\usepackage{graphicx}
\usepackage{textcomp}
\usepackage{array}
\usepackage{bm}
\usepackage{amsmath}
\usepackage{color, colortbl}
\usepackage{xcolor}

\SetCommentSty{mycommfont}

\usepackage[misc]{ifsym}
\usepackage{amsthm}
\newtheorem{Problem definition}{Problem definition}
\newtheorem{Principle}{Principle}
\newtheorem{Example}{Example}

\def\BibTeX{{\rm B\kern-.05em{\sc i\kern-.025em b}\kern-.08em
    T\kern-.1667em\lower.7ex\hbox{E}\kern-.125emX}}

\SetNlSty{}{}{}

\let\oldnl\nl% Store \nl in \oldnl
\newcommand\nonl{%
  \renewcommand{\nl}{\let\nl\oldnl}}% Remove line number for one line
  
\usepackage{hyperref}
\hypersetup{
    colorlinks=true,
    linkcolor=blue,}
\author{
    \IEEEauthorblockN{Yuyan Chen$^{1}$, Yanghua Xiao$^{1,2\,\textrm{\Letter}}$, Bang Liu$^{3,4}$}
    \IEEEauthorblockA{$^1$ Shanghai Key Laboratory of Data Science, School of Computer Science, Fudan University, Shanghai, China}
    \IEEEauthorblockA{$^2$ Fudan-Aishu Cognitive Intelligence Joint Research Center, Shanghai, China}
    \IEEEauthorblockA{$^3$ RALI \& Mila, Université de Montréal, Montréal, Québec, Canada}
    \IEEEauthorblockA{$^4$ Canada CIFAR AI Chair}
    \IEEEauthorblockA{
    \href{mailto:chenyuyan21@m.fudan.edu.cn}{\texttt{chenyuyan21@m.fudan.edu.cn}}, \href{mailto:shawyh@fudan.edu.cn}{\texttt{shawyh@fudan.edu.cn}}, 
    \href{mailto:bang.liu@umontreal.ca}{\texttt{bang.liu@umontreal.ca}}} % 给人名附上邮箱地址
}

\begin{document}

\title{Grow-and-Clip: Informative-yet-Concise Evidence Distillation for Answer Explanation
}

% \and
% \and
% \and
% \and
% \and

\maketitle

\begin{abstract}
    Interpreting the predictions of existing Question Answering (QA) models is critical to many real-world intelligent applications, such as QA systems for healthcare, education, and finance. However, existing 
    % data-driven algorithms and neural network models, such as BERT and GPT3, have achieved big successes over various tasks including QA. However, these 
    QA models lack interpretability and provide no feedback or explanation for end-users to help them understand why a specific prediction is the answer to a question.
    % by a model. 
    In this research, we argue that the evidences of an answer is critical to enhancing the interpretability of QA models.
    % , as well as further improving the performance of QA. 
    Unlike previous research that simply extracts several sentence(s) in the context
    % that contain(s) the answer 
    as evidence, we are the first to explicitly define the concept of evidence as the supporting facts in a context which are informative, concise, and readable.
    Besides, we provide effective strategies to quantitatively measure the informativeness, conciseness and readability of evidence. Furthermore, we propose \textit{Grow-and-Clip} Evidence Distillation (GCED) algorithm to extract evidences from the contexts by trade-off informativeness, conciseness, and readability.
    We conduct extensive experiments on the SQuAD and TriviaQA datasets with several baseline models to evaluate the effect of GCED on interpreting answers to questions. 
    % improving vanilla QA performance, as well as 
    Human evaluation are also carried out to check the quality of distilled evidences. 
    Experimental results show that automatic distilled evidences have human-like informativeness, conciseness and readability, which can enhance the interpretability of the answers to questions.
    % Moreover, if we use ground-truth answers in GCED, GCED-QA models can achieve significant improvement compared with vanilla QA besides interpreting answers. 
    
\end{abstract}


\section{Introduction}
Question Answering (QA) is an important task in Natural Language Processing (NLP) and plays a vital role in many real-world applications, such as search engines \cite{Mitra:20,Shou:20}, chatbots \cite{Avila:20,Jeong:20,Afrae:21}, and so on.
% Recent years have witnessed the super-human performance on QA benchmarks brought by large-scale pre-trained language models, such as BERT \citep{devlin:18} and GPT3 \citep{brown:20}. 
However, most of the current QA systems focus only on extracting or generating the correct answers, but lacking reasonable evidences to support the answers simultaneously. This hurts real users' confidence on the answers and limits the utility of QA systems in many real-world applications that require high interpretability, such as evidence-based medicine \cite{harden:09,sackett:97}, children education \cite{davies:02,weisz:10} and online consulting.
For example, 
% given a question "What is the area of the desert that Ghanzi is in the middle of?". We can use "Kalahari Desert extends for 900000 km2 and Ghanzi is a town in the middle of it." as the evidence to explain the answer "900000 km2". It provides feedback for end-users to show the fact information which explains the answer. 
% Moreover, 
evidences in medical QA \cite{harden:09,sackett:97} offer key clues of patients' symptoms for clinical doctors, assisting them to make correct diagnoses. But treatment without evidences will hardly be trusted or accepted by patients.
% For educational applications \citep{davies:02,weisz:10}, providing evidence or pointing out the key clues to an answer can help students better understand how a difficult problem was solved, easing the burden of teachers. 
% Evidences also play an important role in industry, such as online consulting for customers. 
Therefore, evidences
% , which are usually in the form of natural language,
are important for providing humans suggestive references and important clue information in the real world.

\begin{figure}[!t]
    \centering
    \includegraphics[width=0.95\linewidth]{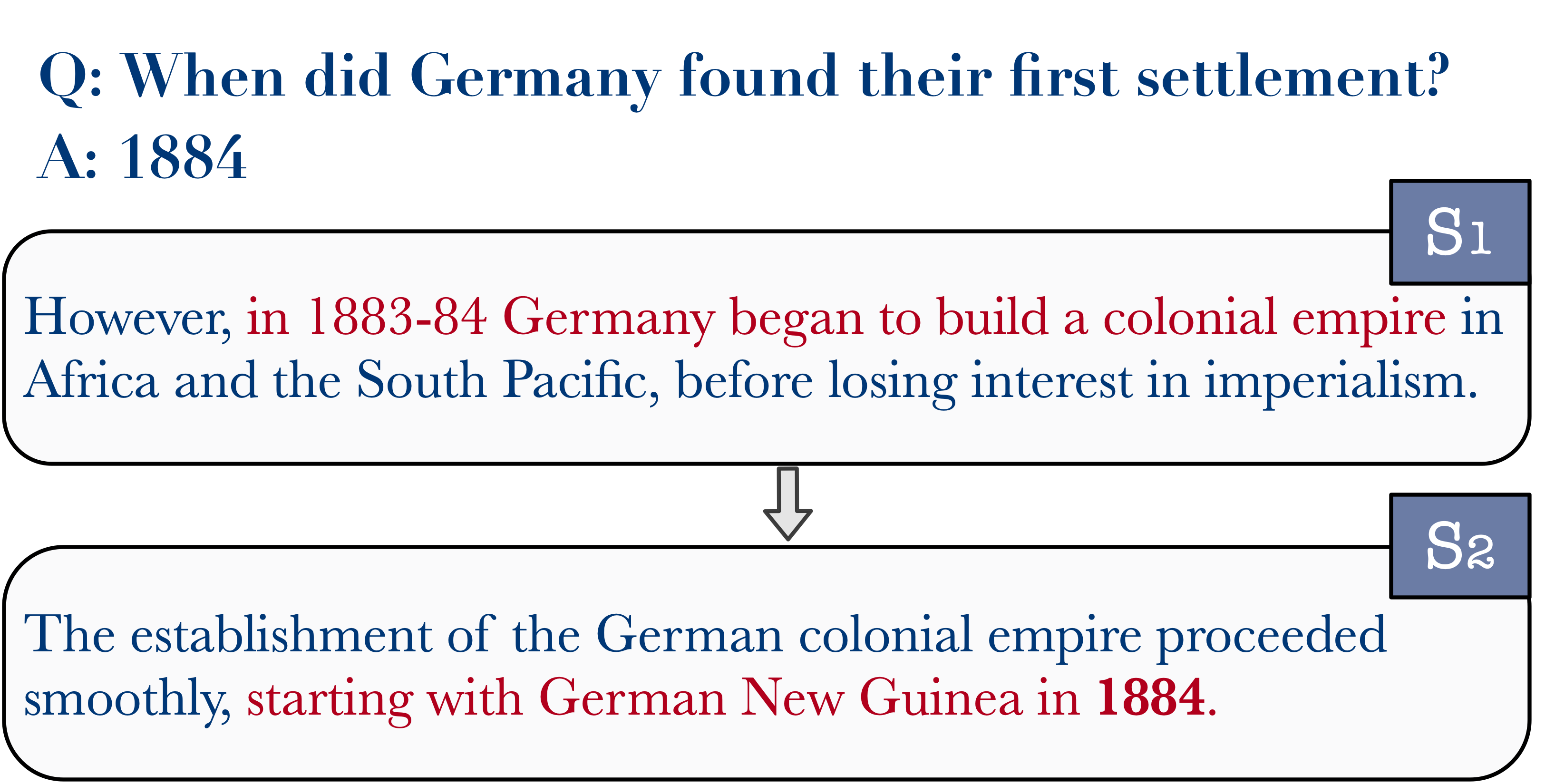}
    \caption{An example of sentence-level evidence for QA. The evidence is composed of two sentences with 37 words. The QA-related part is colored in red with only 16 words, and other parts are considered as noises with 21 words, even longer than the QA-related part.}
    \label{fig:example1}
    \vspace{-4mm}
\end{figure}

%\blue{Problem Statement}
How to define and extract high-quality evidences is a non-trivial problem. A direct idea is taking the documents where the answer comes from as the evidence. But in most cases, a document is too long to be a satisfying evidence. 
% It was shown that a QA model tends to focus on wrong parts of the context when the context is long (e.g., one or multiple documents), leading to incorrect answers~\cite{Jia:17}.
Hence, recent research tends to select a few sentences from factual source text such as Wikipedia~\cite{Chen:17,Yangw:19} or Freebase~\cite{Berant:13, Kwiatkowski:13, Yih:15} as the evidence.
% , which still suffer the noise issue. 
% For a sample QA-pair shown in Fig.~\ref{fig:example1}, the text boxes marked with $S_1$ and $S_2$ in the upper right corner represent the evidence for a given QA pair generated by Min et al.~\cite{min:18}. 
% \texttt{"When did Germany found their first settlement?"} and \texttt{"1884"}, the evidence generated by min et al. \citep{min:18} is \texttt{"However, in 1883-84 Germany began to build a colonial empire in Africa and the South Pacific, before losing interest in imperialism.
% The establishment of the German colonial empire proceeded smoothly, starting with German New Guinea in 1884."},
For a sample QA-pair shown in Fig.~\ref{fig:example1}, $S_1$ and $S_2$ represent the evidence for a given QA pair generated by Min et al.~\cite{min:18}. 
% It has two sentences with 37 words in total.
The QA-related part in this evidence has only 16 words, far shorter than the length of the whole sentence-level evidence (37 words).
% \texttt{"before losing interest in imperialism"} in $S_1$ and \texttt{"proceeded smoothly"} in $S_2$ are irrelevant noises.
% Wang et al.~\cite{{HaiWang:19}} and Wang et al.~\cite{wangsh:18} also generate the evidence at sentence-level or passage-level that entail or support a QA pair from given reference documents, which still suffer the noise issue. 
Therefore, although sentence-level evidences are informative which have a better coverage and contain rich information to support answers, they are still too coarse-grained with irrelevant parts and may introduce noises. 
And noisy information in the evidence will inevitably distract end-users from the most QA-related part of the evidence, increasing the difficulty to understand the explanation and hurting the user-friendliness of the QA system.

% 由此看来，应该删除冗余信息，更加简洁，但是如何删呢？这个怎么度量呢?
Based on the above analysis, we argue that \emph{both informativeness and conciseness are essential for QA evidences.}
However, it is a great challenge to trade-off between informativeness and conciseness of the evidences, as they are contradictory to each other. 
A too informative evidence tends to involve redundant information, while a too concise evidence might miss some relevant information. 
Furthermore, the difficulty is deteriorated by simultaneously guaranteeing readability of the evidences, which is necessary for ensuring user-friendliness of QA systems.
Most previous relevant efforts, such as Schuff et al. \cite{schuff:20}, only focus on informativeness of evidences without considering the conciseness of the evidences.
% from the conciseness aspect. 
Although human experts are further employed to write informative-yet-concise evidences in QA systems~\cite{Rajani:19}, it inevitably
% which however 
incurs unaffordable human cost.

To automatically trade-off between informativeness and conciseness, and guarantee readability of an evidence,
% overcome the above-mentioned challenges, 
we propose a novel task of
\emph{evidence distillation}, which aims at extracting informative-yet-concise evidences from the given contexts to explain answers.
To solve this task, we design a systematical quantitative framework that evaluates the goodness of evidences in terms of \emph{informativeness, conciseness}, and \emph{readability}. Our core idea is that a good evidence shall be supportive to the answer (informativeness), contains few non-essential information to answer the question (conciseness), and easy to understand (readability).
Based on the framework, we further propose a novel Grow-and-Clip Evidence Distillation (GCED) algorithm to distill the optimal evidence while trickily balancing between the informativeness and conciseness. 

Different from previous solutions, GCED finds the optimal evidence at token-level, which is more fine-gained than that at sentence-level and allows more flexibility to generate more informative-yet-concise evidences. 
It is also noteworthy to mention that our GCED algorithm has three advantages compared to previous research:
i) our evidence distillation does not need human annotation; ii) each step is traceable, which enhances the interpretability of generated evidences for QA pairs; and iii) the algorithm is applicable to use either given documents or structured knowledge as the QA source repository, and domain independent.

We carry out human evaluations to systematically assess the quality of our generated evidences from two widely acknowledged QA datasets: SQuAD~\cite{squad:16,squad:18} and TriviaQA~\cite{triviaqa:17}. 
The results verify that our solution is able to find the evidences with high informativeness, conciseness and readability, which can explain answers very well.
% quality and interpretability.
Moreover, if the QA systems have ground-truth answers in an ideal setting or some applications (such as searching engine), 
we can use the ground-truth answers to distill evidences. These evidences are a form of concise and readable contexts which can also explain/support the ground-truth answers. 
% nag

% The empirical results demonstrate that our GCED-QA models significantly outperform vanilla QA models. 

To summarize, our contributions in this paper are threefold:
\begin{itemize}
    \item We establish a quantitative framework to evaluate the goodness (informativeness, conciseness, and readability) of evidences for QA.
    % We also propose human evaluation protocol accordingly, justifying the high quality of our distilled evidences.
    \item We propose a novel Grow-and-Clip Evidence Distillation (GCED) algorithm to distill optimal evidences, with carefully designed heuristic to keep balance between informativeness, conciseness and readability of the distilled evidences. % Specifically, a Grow-and-Clip search strategy over tokens is proposed to find informative-yet-concise and human-readable evidence. 
    %Different from current work which extract evidences at sentence or paragraph level, our framework prominently focus on word level, further filtering out redundant words or phrases in a sentence, acquiring extremely concise but informative and readable evidences.
    \item 
    %We systematically evaluate the QA performance fed with our distilled evidences based on the ground-truth answers.
    % we use the ground-truth answers to distill evidences.
    % when substituting the context with our evidence. 
    %The experimental results show that the ground-truth-based evidences can substitute the contexts to significantly improve the performance of vanilla QA systems besides explaining the answers.
    We systematically evaluate the quality of our distilled evidences based on ground-truth answers and predicted answers with human evaluation. Experimental results demonstrate that the evidences are of good quality (informativeness, conciseness, and readability) in both two situations.
\end{itemize}

\section{Problem formulation}
In this section, we first formulate our problem and introduce the preliminary concepts used in this paper. Then we present the quantitative evaluation framework for QA evidences.

\subsection{Evidence Distillation}
The evidence distillation task in this research can be summarized as the following: given a tuple set $S=\{(q_i, a_i, c_i, e_i)\}_{i=1}^N$ where $N$ refers to the total number of QA pairs, evidence distillation aims to take the natural language question $q_i$, the predicted answer $a_i$ as well as the given contexts $c_i$ as the input, and outputs a good evidence $e_i$ which helps to explain why $a_i$ is the answer to the question $q_i$. 
Compared to the full context $c_i$, the distilled evidence $e_i$ is more concise but preserves essential information for explaining the answer $a_i$.
%输入输出定义
% \begin{problem definition}
% This task accepts $q_i$, $a_i$ and $c_i$ (representing the question, the ground-truth answer and the context, respectively) as the input, and generates $e_i$ as the output, where $e_i$ is the distilled optimal evidence for a QA pair. 
% \end{problem definition}

\begin{figure}[!t]
    \centering
    \includegraphics[width=0.95\linewidth]{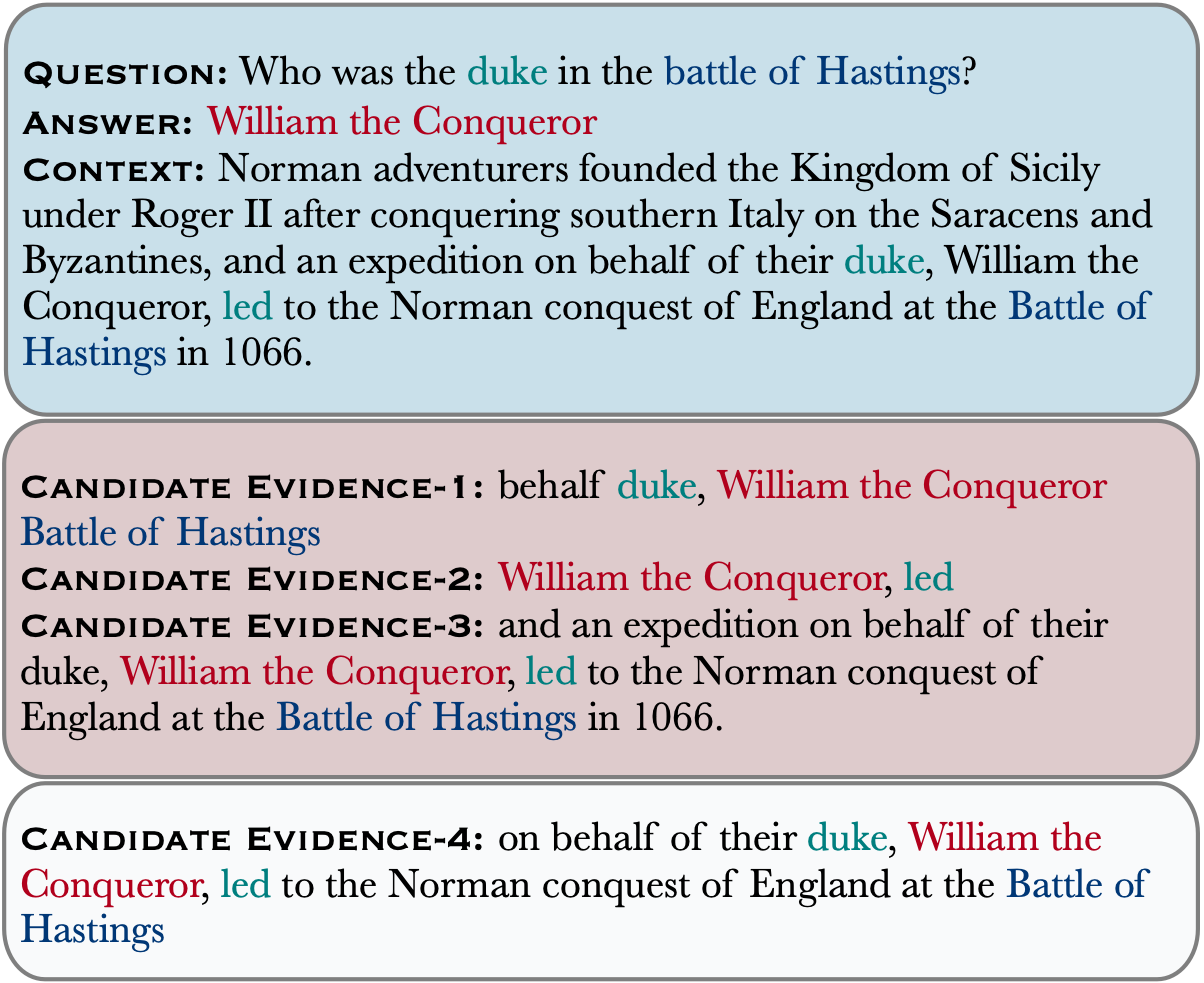}
    \caption{An example of four candidate evidences for the given QA pair. The first three are informative or readable or concise. The fourth satisfies all of the three characteristics (informative and readable and concise), so it's considered as a good evidence.}
    \label{fig:example2}
    \vspace{-4mm}
\end{figure}

A critical research problem in our task is how to evaluate the goodness of an evidence.
We argue that a good evidence is expected to have the following three characteristics:
% Here, we display the characteristics of a good evidence and its quantitative evaluation as follows:
\begin{itemize}
    \item i) \emph{Informativeness}. A good evidence is informative so that the input answer can be inferred from the evidence.
    % Language models (LMs) are employed to substitute evidences for contexts to establish QA models. F1-scores of the models quantify the informativeness of evidences; 
    For example, the first candidate evidence shown in Fig.~\ref{fig:example2} contains more relevant information to infer the answer of this question, such as \texttt{"duke"} and \texttt{"Battle of Hastings"}. It is thus more informative.
    On the contrary, the second candidate evidence only contains limited information such as \texttt{"William the Conqueror"} and \texttt{"led"}, so we cannot infer the answer from this evidence. Therefore, it's not informative.
    %正向举例informative
    \item ii) \emph{Conciseness}. A good evidence is expected to be concise which is as short as possible but at least longer than the answer.
    % so that deleting any words will lead to another answer.
    % Therefore, the reciprocal of the length of evidences is used to measure their conciseness.
    For example, 
    the second candidate evidence shown in Fig.~\ref{fig:example2} only has 4 words, just one more word than the answer which have 3 words. Therefore, it's considered concise.
    On the contrary, the third candidate evidence is quite redundant with 25 words, much longer than the answer, which is not concise.
    % We can also find that it contains redundant information, such as \texttt{"in 1066"}, which is not essential to the answer.
    \item iii) \emph{Readability}. A good evidence is readable and and (almost) grammatically correct, so that it is understandable by humans.
    % Perplexity of evidences and its corresponding full sentence in contexts are calculated. The ratio between them indicates the readability of evidences; 
    For example, the third candidate evidence shown in Fig.~\ref{fig:example2} has clear logic and doesn't have grammar mistakes. The meaning is easy to understand, and thus is considered readable.
    On the contrary, the first candidate evidence
    % contains more relevant information to answer the question, such as \texttt{"duke"} and \texttt{"Battle of Hastings"}, so it is more informative. However, it 
    has grammar mistakes, such as missing a preposition between \texttt{"behalf"} and \texttt{"duke"}, and missing a verb between \texttt{"William the Conqueror"} and \texttt{"Battle of Hastings"}. Therefore, it's difficult to understand with bad readability.
\end{itemize}

Compared with the three evidences, the fourth candidate evidence shown in Fig.~\ref{fig:example2} satisfies all the three characteristics.
It contains enough information to answer the question without redundant noises. It also has better readability without grammar mistakes, which is easy to understand for humans.
Therefore, the fourth candidate evidence is considered as a good evidence.
%Delete any word in it may lead to a wrong answer. Therefore, it's considered as a good evidence.

\subsection{Metrics of a Good Evidence}
\label{Metrics of A Good Evidence}
Next, we quantify the three above-mentioned characteristics in order to measure the goodness of an evidence. 
% We leverage machine evaluation to distill optimal evidence and adopt human evaluation to rate the distilled evidences.
% We mathematically elaborate criteria of evidence distillation based on three characteristics of evidence: informativeness, readability and conciseness.
\subsubsection{Informativeness}
\label{Informativeness}
The first important characteristic of an evidence is informativeness. 
We resort to a QA model, such as a large-scale pre-trained language model (denoted as $PLM$), to evaluate the informativeness of an evidence.
\begin{Principle}
An evidence is informative if a given answer can be predicted from this evidence by a QA model.
\end{Principle}
% The basic principle is that an evidence is informative if the ground-truth answer is likely to be predicted from the evidence by a good QA model, e.g., a large-scale pre-trained language model. 
% % (denoted by $LM$). 
The evidence is more informative if the input answer is predicted more accurately from the evidence by the QA model.
$PLM$ is a statistical model that has already been trained by mountains of data, and has been widely acknowledged as the foundation of variant NLP tasks, such as reading comprehension, dialog, etc. 
In this paper, we use large-RoBERTa~\cite{RoBERTa} as $PLM$, which can be easily replaced with other pre-trained language models (such as BERT~\cite{devlin:18}, etc) without significant influence on the results. The detailed process is as follows:
\begin{itemize}
    \item Step 1 (\emph{\bf QA model training}): In this step, we train a QA model based on $PLM$. 
    % We first use $LM$ to encode the question and context. 
    We first separate the question $q_i$ and the context $c_i$ by a special separator \texttt{[SEP]}, then divide the context $c_i$ into several segments with a sliding window to keep the most informative context segment. 
    Next, we use \texttt{"padding"} to make the length of the input sequence the same. 
    After that, based on the encoded question and context, we train a QA model aiming at minimizing the training loss. 
    % Kullback-Leibler divergence loss. % \red{$\text{KL}(Y||P)$}
    %参考文献
    \item Step 2 (\emph{\bf Answer prediction}): In this step, we input the question $q_i$ and the evidence $e_i$ into the QA model trained in the above step to predict a new answer $\hat{a_i}$.
    %, as shown in Eq.~\ref{eq:ft}.
    % We replace the original context $c_i$ with the evidence $e_i$, feeding the evidence $e_i$ and the question $q_i$ into the QA model which is trained in the last step, to predict the answer $\hat{a_i}$ shown in Eq.~\ref{eq:ft}. 
    Specifically, answer prediction is based on the maximum probability calculated by a Softmax function, and tokens of the highest two probabilities are taken as the start and the end of $\hat{a_i}$.
    \item Step 3 (\emph{\bf Evidence informativeness evaluation}): In this step, we use the predicted answer and the input answer to evaluate the informativeness of the evidence. 
    Inspired by Schuff et al.\cite{schuff:20}, we use the overlap between the predicted answer $\hat{a_i}$ and the input answer $a_i$ to evaluate the informativeness of the evidence $e_i$. 
    Specifically, we use F1 score, which is widely used to evaluate the accuracy of answer prediction in machine reading comprehension~\cite{squad:16}, as the informativeness score $I(e_i)$ of the evidence $e_i$. The procedure is shown in Eq.~\ref{eq:infor}:
    \begin{gather}
\label{eq:infor}
\small
\begin{aligned}
    &Pre(e_i)=\frac{N_c(\hat{a_i},a_i)}{L(\hat{a_i})},
    Rec(e_i)=\frac{N_c(\hat{a_i},a_i)}{L(a_i)}\\
    &I(e_i)=\frac{2\times Pre(e_i)\times Rec(e_i)}{Pre(e_i)+Rec(e_i)}
\end{aligned}
\end{gather}
\noindent where $N_C$ means the number of common tokens in two sequences, and $L$ means the length of a sequence. 
\end{itemize}
% \begin{gather}
% \label{eq:ori}
%     \dot{a_i}=LM(q_i,c_i)
% \end{gather}
% \begin{gather}
% \label{eq:ft}
% \small
% \end{gather}

\subsubsection{Conciseness} 
The second important characteristic of an evidence is conciseness. We directly use the reciprocal of the length of the evidence to measure its conciseness.

\begin{Principle}
An evidence is concise if it is as short as possible but at least longer than the answer.
% if deleting any words in the evidence will lead to another answer.
\end{Principle}
We introduce a conciseness score defined in Eq.~\ref{eq:con} to quantify the conciseness of an evidence.  
The longer of the evidence, the lower conciseness score of it. 
If the length of the evidence is no longer than the length of the answer, we discard this evidence. 
% Specifically, the conciseness of an evidence is quantified as Eq.~\ref{eq:con}:
% \begin{gather}
% \label{eq:con}
%     C(e_i)=\frac{1}{L(e_i)} (L(e_i)>L(a_i))
% \end{gather}
\begin{equation}
\label{eq:con}
C(e_i)=\left\{
\begin{array}{rcl}
\frac{1}{L(e_i)} & & {L(e_i)>L(a_i)}\\
-\infty & & {L(e_i)\leq L(a_i)}
\end{array} \right.
\end{equation}

\subsubsection{Readability}
The third important characteristic of an evidence is readability. 
%Because the process of trading-off between informativeness and conciseness of an evidence may hurt its readability,
% too much information in an evidence will make it redundant and too concise of the evidence will miss some clue information, 
%We introduce the readability score to evaluate the readability of an evidence. 
Specifically, we use the reciprocal of the perplexity as the readability score. The perplexity of an evidence is based on the QA model trained in the informativeness evaluation. The higher the readability score is, the better readability of the evidence. 
% The score is calculated with two steps. In the first step, 
The calculating process of the readability score is shown in Eq.~\ref{eq:ppl}:
% is further normalized by Eq.~\ref{eq:normalize}, where we use the ratio of evidence perplexity to answer-oriented sentence(s) perplexity as the normalized score:
\begin{equation}
\small
\label{eq:ppl}
    \begin{aligned}
        PPL(e_i)&=P({(e_i)}_1,{(e_i)}_2,\cdots,{(e_i)}_L)^{\frac{1}{L}}\\
        &=\sqrt[L]{\prod_{j=1}^{L}\frac{1}{p({(e_i)}_j|{(e_i)}_1,{(e_i)}_2,\cdots,{(e_i)}_{j-1})}}
    \end{aligned}
\end{equation}
\begin{gather}
% \label{eq:normalize}
\small
    R(e_i)=\frac{1}{PPL(e_i)}
\end{gather}
\noindent where $P(e_i)$ and $PPL(e_i)$ are the probability of generating the evidence $e_i$ by the QA model and the perplexity of the evidence $e_i$, respectively.
% $T$ is the length of $e_i$,
% $e({(x_i)}_t)$ represents the embedding of word ${(x_i)}_t$, $\hat{x_i}$ and $\bar{x_i}$ represent unmasked and masked words, respectively. $m_t=1$ represents \verb"MASK" at time $t$ which needs recovery, $H_\theta({x_i})=[H_\theta{(x_i)}_1,H_\theta{(x_i)}_2,\cdots,H_\theta{(x_i)}_T]$ represents a sequence of hidden state mapped by $x_i$ with length $T$.
% $PPL(e_i)$ and $PPL(s_i)$ are perplexity of $e_i$ and the answer-oriented sentence(s) $s_i$ (which will be elaborated in Sec.~\ref{ASE}), respectively.

\subsubsection{Hybrid score}
\label{Hybrid score}
For the sake of balancing informativeness, conciseness and readability, we set three weight factors $\alpha$, $\beta$ and $\gamma$ to derive a Hybrid score $H(e_i)$ for the evidence $e_i$: 
% Hybrid scores will be used to determine the optimal path of Grow-and-Clip strategy.
\begin{gather}
\label{eq:hs}
\small
    H(e_i)=\alpha I(e_i)+\beta R(e_i)+\gamma C(e_i)
\end{gather}
\noindent where 
% $I$, $R$, and $C$ are introduces in Sec.~\ref{Metrics of A Good Evidence}. 
$\alpha>0$, $\beta>0$ and $\gamma>0$, $\alpha+\beta+\gamma=1$. The detailed value of $\alpha$, $\beta$ and $\gamma$ will be determined by experiments. $H(e_i)=0$ means $e_i$ is a worst evidence while $H(e_i)=1$ means $e_i$ is a best evidence.

\section{Methods}
\label{Methods}
In this section, we introduce the Grow-and-Clip Evidence Distillation (GCED) framework and elaborate its major modules.%: Answer-oriented Sentences Extractor, Query-based Words Selector, Weighted Syntactic Parsing Tree Constructor, Evidence Forest Constructor and the Optimal Evidence Distiller.

\subsection{Algorithm Framework}
\begin{figure*}[!t]
    \centering
    \includegraphics[width=0.95\linewidth]{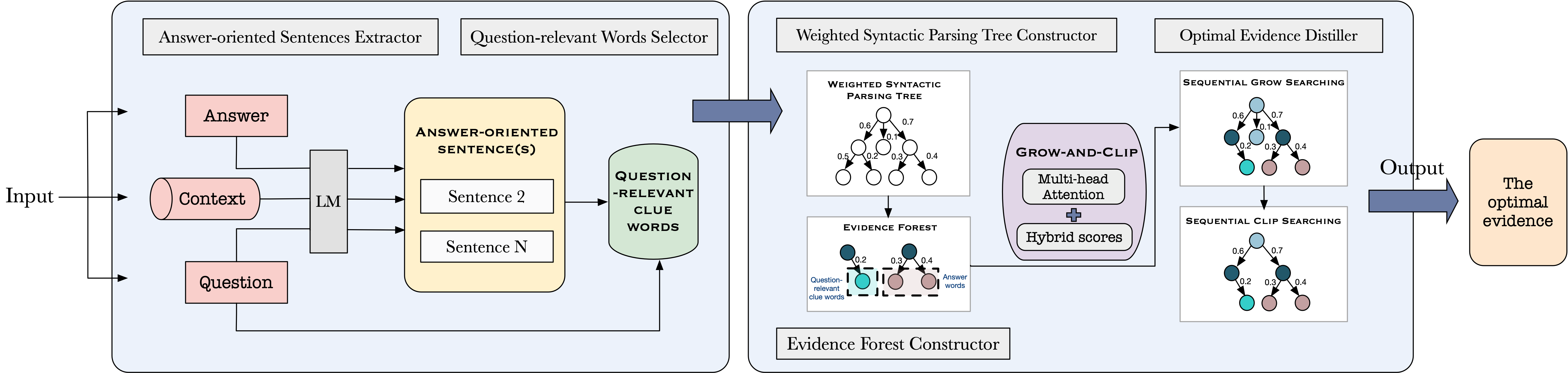}
    \caption{Framework of proposed GCED with five modules: the Answer-oriented Sentences Extractor, the Query-based Words Selector, the Weighted Syntactic Parsing Tree Constructor, the Evidence Forest Constructor and the Optimal Evidence Distiller.}
    \label{fig:framework}
\end{figure*}

The overall framework of GCED is displayed in Fig.~\ref{fig:framework}.
It consists of five core modules. Answer-oriented Sentences Extractor extracts sentences that are semantically related to a QA pair from the context. The sentences are referred as answer-oriented sentence(s). Question-relevant Words Selector selects question-relevant clue words in the answer-oriented sentences.
Weighted Syntactic Parsing Tree Constructor adopts Lexicalized Probabilistic Context-Free Grammars (L-PCFGs) and multi-head attention to construct a weighted syntactic parsing tree for the answer-oriented sentences.
Evidence Forest Constructor adopts dependency parser to discover evidence segments from the weighted syntactic parsing tree. Optimal Evidence Distiller adopts Grow-and-Clip strategy based on the hybrid scores and attention weights to distill a good evidence.

\subsection{Answer-oriented Sentences Extractor (ASE)}
\label{ASE}
\begin{figure}[!t]
    \centering
    \includegraphics[width=0.95\linewidth]{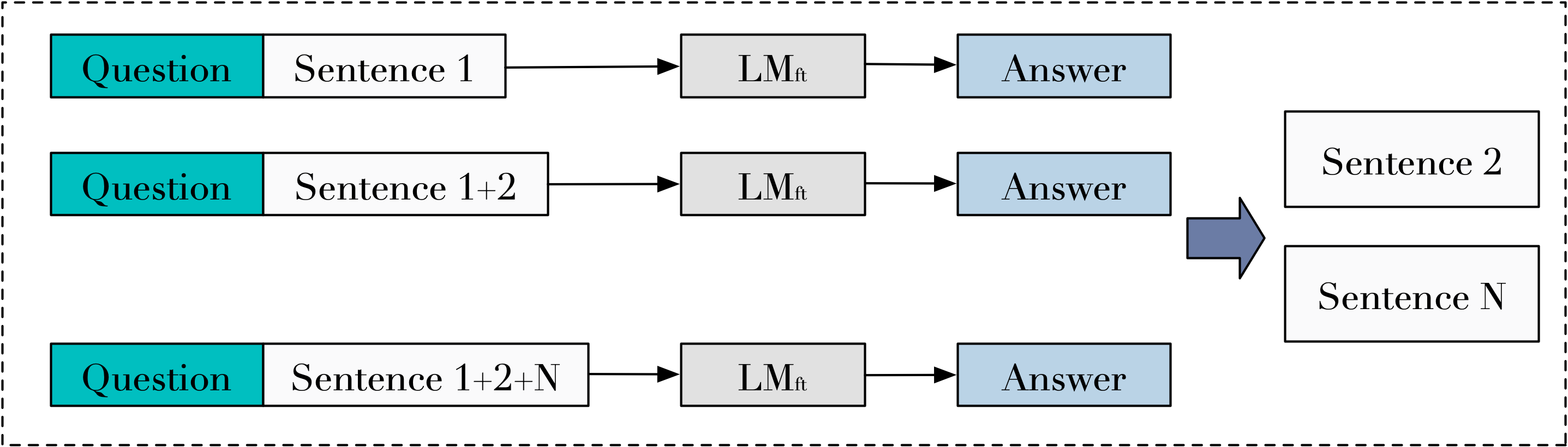}
    \caption{Framework of ASE. 
    % Subfigure (a) represents the probability of a QA pair semantic correlated with each sentence. Subfigure (b)
    It represents the answers predicted by different sentence subsets. Assume there are $N$ sentences in a context, we select sentence 2 and sentence $N$ from it based on a QA model.}
    \label{fig:sentence-selector}
\end{figure}

%In this part, we build an Answer-oriented Sentences Extractor (ASE) to extract the answer-oriented sentence(s). 
Answer-oriented sentence(s) are the minimum sentence subset that has enough information to predict a given answer. The minimum sentence subset may contain one or several sentences.
The input of this module is the QA pair and one or more sentences in the context, and the output is the answer-oriented sentence(s). 
We 
% % rank all sentences in the context based on the semantic relatedness between a QA pair and each sentence. Then we 
use a QA model to select sentences which are able to predict (that is most semantically related one) given answer to obtain the answer-oriented sentence(s).
% we put each sentence in the context into the QA model to select the sentences which are able to predict (the closet) given answer. The minimum sentence subset which can predict (the closet) given answer is called answer-oriented sentence(s). 
Specifically, we first feed the sentences in the context one at a time 
% based on their probability (calculated in the Step 1) in the descending order 
into the QA model to make answer prediction. 
% The context is substitute by sentences in the candidate pool to combine with the question to make answer prediction.
After that, the QA model stops accepting sentences when it predicts the input answer for the first time.
Sentences so far are regarded as the minimum sentence subset that covers enough information, and are expected to predict the input answer. These sentences are called the answer-oriented sentence(s). 
If the QA model fails to predict the input answer after processing all sentences, we calculate the overlap between the input answer $a_i$ and each predicted answer $\tilde{a_i}$ by Eq.~\ref{eq:infor}. After that, the sentence subset with the maximum overlap is taken as the answer-oriented sentence(s).
The detailed process is shown in Fig.~\ref{fig:sentence-selector}.

% Each sentence in the context is scored 0 or 1 according to Eq.~\ref{answer-oriented}, and we extract the sentence scored 1 as the answer-oriented sentence $s_i$.
% \begin{equation}
%   \label{answer-oriented}
%   s_i=
%   \begin{cases}
%   1& \text{if sentence $s_i$ contains answer $a_i$}\\
%   0& \text{otherwise}
%   \end{cases}
% \end{equation}

% The discrete random variable $X=\{x_{ij}\} (i=1,2,\cdots, n;j=1,2,\cdots,k)$ obeys 0-1 distribution shown in Eq.~\ref{answer-oriented}, where $x_{ij}=1$ represents the sentence $s_i\in \{s_1,s_2,\cdots,s_o\}$ which comes from the context $c_i$ contains the input answer $a_i$ and $x_{ij}=0$ means the opposite. $n$ and $m$ signify the number of answers, contexts in the dataset, and $o$ denotes sentences in the corresponding context. 

% Furthermore, words in the answer-oriented sentence $s_i$ are encoded by position whose indexes are $p_j^{s_i}(j=1,2,\cdots,m)$ where $m$ represents the length of the sentence $s_i$. We adopt the same operation for the corresponding answer $a_i$ to generate indexes $p_j^{a_i}(j=1,2,\cdots,n)$ where $n$ represents the length of the answer $a_i$.

\subsection{Question-relevant Words Selector (QWS)}
\label{QWS}
This module select the question-relevant clue words from the answer-oriented sentence(s). The words in the answer-oriented sentence(s) that are semantically relevant with the significant words in the question are regarded as question-relevant clue words.
Specifically, we first remove insignificant words in the question. Insignificant words include all question terms (such as \textit{who, where}), auxiliary verbs (such as \textit{do, did}), functional words (such as \textit{conj, art, prep, pron}) and punctuations such as \textit{!?,.()}.
Next, for any remaining word in the question, if the word and its synonyms, antonyms, sibling terms sharing the same hypernym (by lookup from WordNet) appear in the answer-oriented sentence(s), these words are regarded as the question-relevant clue words~\cite{fellbaum:10}.  
% \end{Principle}

%The searching breadth is set as 30.

% Given question $q_i$ and answer-oriented sentence $s_i$, we select out clue words in $s_i$ which contains $w^{s_i}=\{w^{s_i}_1,w^{s_i}_2,\cdots,w^{s_i}_j,\cdots,w^{s_i}_{k}\}$ on the basis of $q_i$ which includes $w^{q_i}=\{w^{q_i}_1,w^{q_i}_2,\cdots,w^{q_i}_j,\cdots,w^{q_i}_d\}$. Here, $d$ and $k$ describe number of selected tokens in $q_i$ and $s_i$, respectively.

\begin{figure}[!t]
    \centering
    \includegraphics[width=0.95\linewidth]{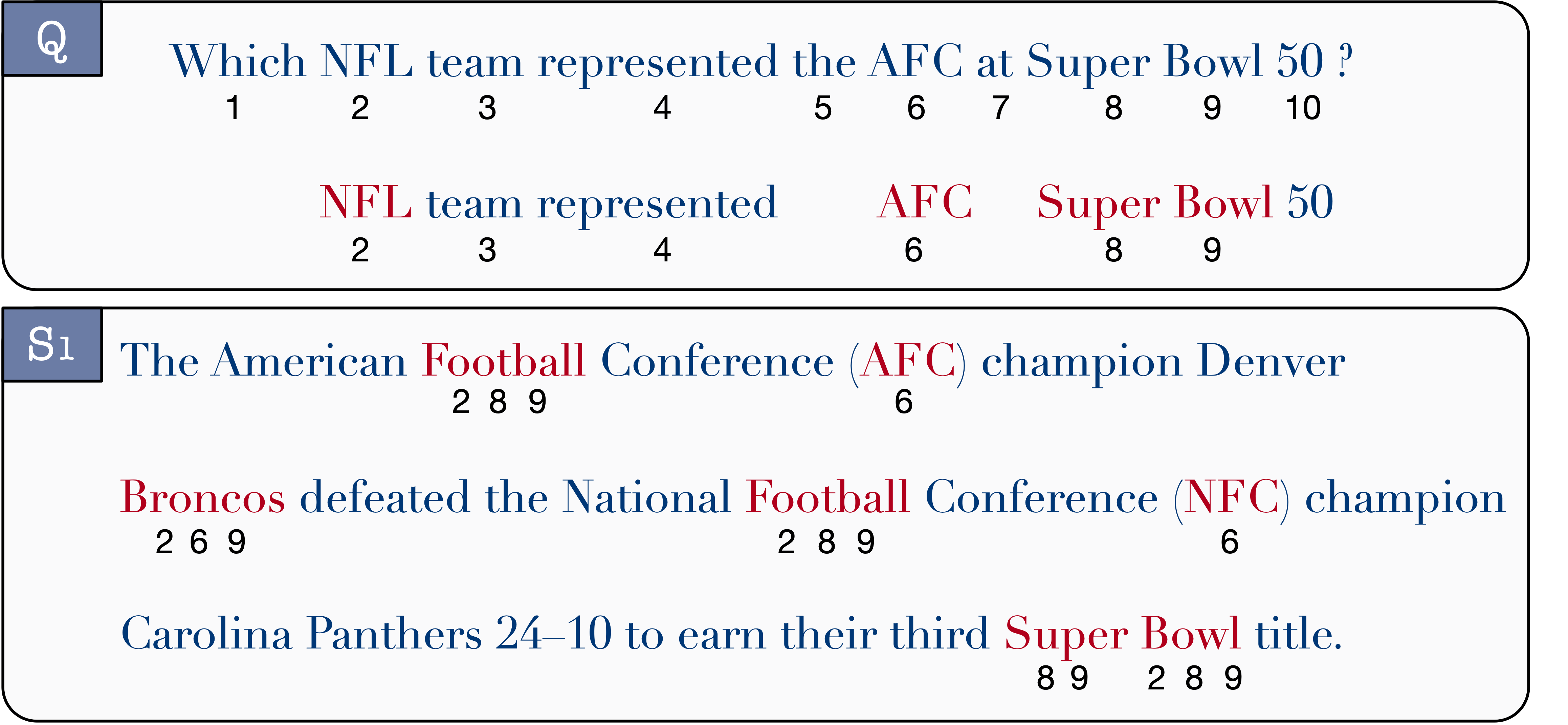}
    \caption{A sample of selecting question-relevant clue words by QWS. The first line in $Q$ is a question, and the second line is remaining words after removing insignificant words. $S_1$ is the answer-oriented sentence(s). Words colored in red represent related-word-pairs in $Q$ and in $S_1$. Numbers below words in $Q$ represent their index, and numbers below words in $S_1$ represent the corresponding words in $Q$.}
    \label{fig:word-selector}
    \vspace{-4mm}
\end{figure}

We use the example in Fig.~\ref{fig:word-selector}) to elaborate the process. The sample question is \texttt{"Which NFL team represented the AFC at Super Bowl 50?"}. From the significant words (such as "NFL"), we find question-relevant clue words \texttt{"Football"}, \texttt{"AFC"}, \texttt{"Broncos"}, \texttt{"NFC"}, \texttt{"Super"}, and \texttt{"Bowl"} in the answer-oriented sentence(s).

\subsection{Weighted Syntactic Parsing Tree Constructor (WSPTC)}
\label{attention_weights}
This module establishes a relationship among all words in the answer-oriented sentence(s) with a tree structure. We first use Lexicalized Probabilistic Context-Free Grammars (L-PCFGs) $Gr=(M,\sum, R, S)$~\cite{collins:99, charniak:05} to generate a syntactic parsing tree.
Each node in the tree has an index, which represents the position of this word in the answer-oriented sentence(s).
% of the answer-oriented sentence(s). 
Then we associate an attention weight to each edge in the syntactic parsing tree. We use the attention derived from the first-layer encoder of $PLM$ as the attention weights. Specifically, we first derive embeddings of each word in the answer-oriented sentence(s) from $PLM$, denoted as $\vec{x^i}$. Then, we adopt linear transformation to obtain representation of queries $Q$, keys $K$ and values $V$ shown in Eq.~\ref{eq:QKV}. After that, we make scaled dot-product attention for $heads=16$ times and calculate self-attention score of each attention head for each vector pair. We make scale by $d_k=64$ and normalize with softmax for $Q$ and $K$ shown in Eq.~\ref{eq:head}. Finally, we concatenate outputs of all attention heads to obtain the attention weights $W^{Attn}_i$ shown in Eq.~\ref{eq:multihead}. $\bm{W_t^Q}$, $\bm{W_t^K}$, $\bm{W_t^V}$, $\bm{W_o}$ are all trainable parameters.

After that, we annotate the attention weights for all edges of the syntactic parsing tree to build a weighted syntactic parsing tree shown in Fig.~\ref{fig:grow-and-clip} (a).
Higher weights between two nodes represent more attention from a node to its child node in the answer-oriented sentence(s). 
\begin{gather}
\label{eq:QKV}
\small
    Q=\bm{W_q}\vec{x^i},K=\bm{W_k}\vec{x^i},V=\bm{W_v}\vec{x^i}
\end{gather}
\begin{gather}
\label{eq:head}
\small
    \begin{aligned}
    &Q_t^i=Q\bm{W_t^Q},K_t^i=K\bm{W_t^K},V_t^i=V\bm{W_t^V},1\leq t\leq16\\
    &head_t^i=Attention(Q_t^i,K_t^i,V_t^i)=softmax(\frac{{Q_t^iK_t^i}^T}{\sqrt{d_k}})V_t^i
    \end{aligned}
\end{gather}
\begin{gather}
\label{eq:multihead}
\small
    \begin{aligned}
    \bm{W^{Attn}_i}&=MultiHead(Q^i,K^i,V^i)\\
    &=Concat(head_1^i,\cdots,head_T^i)\bm{W_o}
	\end{aligned}
\end{gather}

\subsection{Evidence Forest Constructor (EFC)}
\label{Dependency Forest}
%Lexicalized PCFGs 构建句法解析树，word selsctor的每个在answer-oriented sentence中的词构建依赖森林，加上答案这个子树
In this part, we use question-relevant clue words, the input answer and the weighted syntactic parsing tree ($\mathcal{T}$) to construct the evidence forest ($\mathcal{T}^{E}$). 
We find the question-relevant clue words and answer words from $\mathcal{T}$, and the subtrees induced from these words as well as their parents in $\mathcal{T}$ constitute $\mathcal{T}^{E}$. For example, in Fig.~\ref{fig:grow-and-clip} (b), nodes 3, 5, 7 are question-relevant clue words, nodes 13 and 15 are answer words. There parents as well as these nodes constitute two evidence trees (tree with nodes 2, 3, and tree with nodes 5, 6, 7) and one answer tree (tree with nodes 13, 14 and 15). These trees compose of the final evidence forest.

\subsection{Optimal Evidence Distiller (OEC)}
\label{Optimal Evidence Distiller}
\begin{figure*}[!t]
    \centering
    \includegraphics[width=0.95\linewidth]{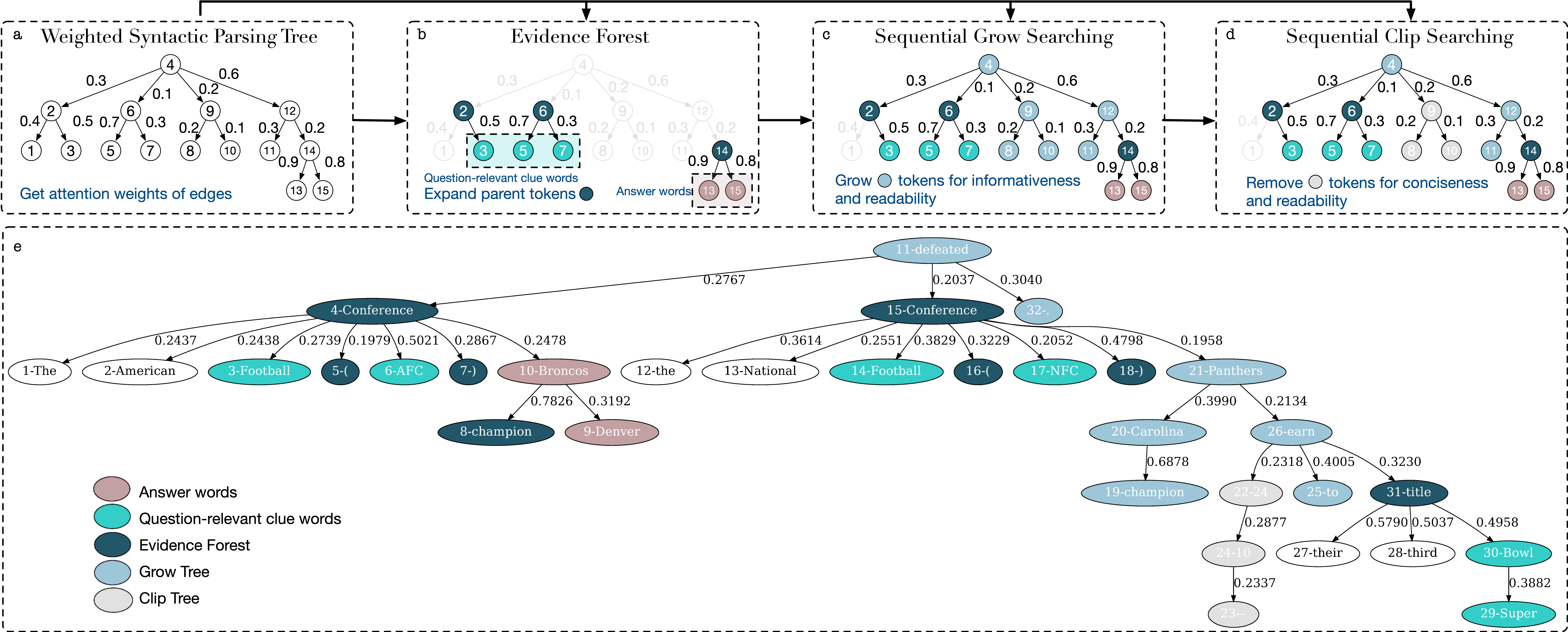}
    \caption{Subfigure (a) to (d) represent the process of the Grow-and-Clip strategy. We use the weighted syntactic parsing tree and the evidence forest to make SGS and SCS so as to distill the optimal evidence. Subfigure (e) displays a detailed process of distilling the optimal evidence by GCED.}
    \label{fig:grow-and-clip}
\end{figure*}

Finally, we distill the optimal evidence from the evidence forest. The evidence forest is still not necessarily to be informative or readable. OEC is designed to address this issue, that is keeping a balance between informativeness. conciseness and readability of the evidences. The input of this module is $\mathcal{T}$ and the evidence forest $\mathcal{F}^{E}$, and the output is the optimal distilled evidence which is the final output evidence. OEC adopts a Grow-and-Clip strategy to distill the optimal evidence based on hybrid scores and attention weights. 
%to distill the evidence with the maximum hybrid score.
%It can resist the alteration of correct answers when substituting the distilled evidence with the context. It also has fluency constraint and length penalty for the distilled evidence. 
% display a heatmap shown in Fig.~\ref{fig:Attention-weights} to visually describe attention-based weighted path. Deep color represents more attention between two words. Take words in rows as roots, and words in columns as their corresponding children, we annotate attention weights for all edges in the syntactic parsing tree.
%
% \begin{figure}[!t]
%     \centering
%     \includegraphics[width=\linewidth]{Attention-weights.png}
%     \caption{Attention-based weighted path in the syntactic parsing tree to search Shortest Path Routing in evidence forest. Deep color represents more attention between words in the row and their corresponding column, which has more attention weights between these word-pairs.}
%     \label{fig:Attention-weights}
% \end{figure}
% \begin{figure*}
%   \centering
%   \includegraphics[width=\linewidth]{tree.png}
%   \caption{A Sample of a weighted evidence tree constructed by Grow-and-Clip strategy to distill optimal evidence. The evidence is characterised by informativeness, readability and conciseness.}
%   \label{fig:tree}
% \end{figure*}
% \subsubsection{Grow-and-Clip}
The Grow-and-Clip strategy searches for a shortest path in $\mathcal{T}$ to connect evidence trees and the answer trees in the evidence forest (introduced in Sec.~\ref{Dependency Forest}). 
It consists of two fundamental search operations: \textit{Sequential Grow Searching (SGS)} and \textit{Sequential Clip Searching (SCS)}, as illustrated in Fig.~\ref{fig:grow-and-clip}(c), (d).
The algorithm of Grow-and-Clip search strategy is illustrated in Algorithm~\ref{algorithm1}.

% \textbf{SGS} merges all trees based on the decreasing attention weights to guarantee \emph{informativeness} and \emph{readability}.
% \textbf{SCS} prunes some leaves based on the decreasing hybrid scores (see Eq.~\ref{eq:hs}) and increasing attention weights (see Eq.~\ref{eq:multihead}) to ensure \emph{conciseness} and \emph{readability}.

%补充这段的过程，伪代码仔细检查
% We construct a weighted syntactic parsing tree with question-relevant clue words and answer words (Fig.~\ref{fig:grow-and-clip}(a)). 
% Then, we build the weighted evidence forest with evidence subtrees (Fig.~\ref{fig:grow-and-clip}(b)). Each evidence subtree consists of question-relevant clue words (or answer words) and their corresponding parent nodes.
\subsubsection{Sequential Growth Searching (SGS)}
We use SGS to connect all trees in the evidence forest ($\mathcal{F}^{E}$) to obtain an informative and readable evidence (Fig.~\ref{fig:grow-and-clip} (c)). SGS achieves this goal by a step-wise growth of a carefully selected optimal tree in $\mathcal{F}^{E}$. In each iterative step, a parent node that has the maximal weight with the roots in the current evidence forest is selected to expand. In this way, we ensure the final evidence evidence is \textit{informative} and \textit{readable}.
The detailed procedure of SGS is as follows and illustrated in Alg.~\ref{algorithm1} and Example~\ref{grow}.
\begin{itemize}
    \item Step 1 (\textbf{Select an optimal tree}): We first select the tree whose root has a maximum weight $W^{attn}$ with its parent node from the evidence forest $\mathcal{F}^{E}$ (shown in line 3 of Grow Step in Alg.~\ref{algorithm1}). This tree is denoted as $\mathcal{T}_{opt}$. $W^{attn}$ is calculated in Sec.~\ref{attention_weights}.
    A higher attention weight means more dependence between the root node and its parent node.
    % Therefore, it may hurt the readability of the evidence if only remaining this node and discarding its parent node.
    \item Step 2 (\textbf{Update the evidence forest}): 
    % Let $r$ be the parent of the root of $\mathcal{T}_{opt}$. 
    Next, we merge $\mathcal{T}_{opt}$ with its parent node and sibling subtrees to replace $\mathcal{T}_{opt}$ (shown in line 4 of Grow Step in Alg.~\ref{algorithm1}). Note that this step will update the evidence forest. This step ensures the readability as much as possible since the new $\mathcal{T}_{opt}$ only expands to nodes that are closely-related to the old $\mathcal{T}_{opt}$ (i.e., parent node and sibling nodes of the old root).
    \item Step 3 (\textbf{Construct an unclipped evidence tree}): We perform Step 1 and 2 repeatedly on the newly updated $\mathcal{F}^{E}$ until all trees $\mathcal{F}^{E}$ are connected. Then an unclipped evidence tree $\mathcal{T}^{E}$ is constructed (shown in line 7 of Grow Step in Alg.~\ref{algorithm1}).
    This unclipped evidence tree is an informative and readable evidence after ranking its nodes by indexes.
\end{itemize}

\begin{Example}[SGS]
\label{grow}
\small
In this example (Fig.~\ref{fig:grow-and-clip} (e)), there are two evidence trees ($T_1$ and $T_2$), and one answer tree ($T_3$). The question-relevant clue words are node 3, 6, 14, 17, 29, 30 (colored in Tiffany blue) and the input answer words are node 9,10 (colored in Sakura pink). %Other words in the evidence forest are node 4, 5, 7, 8, 15, 16, 18, 31 (colored in dark green). 
\begin{itemize}
    \item We first select an optimal tree whose root has a maximum attention weight with its parent node.
    There are three roots nodes in the evidence forest: \texttt{"4-Conference"}, \texttt{"15-Conference"} and \texttt{"31-title"}. Their attention weights with their corresponding parent nodes are 0.2767, 0.2037 and 0.3230, respectively. Therefore, $T_3$ rooted at node 31 is selected as $\mathcal{T}_{opt}$ (shown in Grow Step line 3 in Alg.~\ref{algorithm1}).
    \item Then, we select the parent of node 31 to expand. That is \texttt{"26-earn"}. As a result $\mathcal{T}_{opt}$ is updated to be the subtree rooted at node 26. The new $\mathcal{T}_{opt}$ as well as $T_1, T_2$ is new evidence forest (shown in Grow Step line 4 in Alg.~\ref{algorithm1}). 
    % Then, we repeat the above steps.
    %We combine the parent node \texttt{"26-earn"} and the other children nodes of this parent node \texttt{"22-24"}, \texttt{"24-10"}, \texttt{"23--"}, \texttt{"25-to"} with the selected tree (tree with node 31, 30, 29) to construct a new tree (tree with node 26, 22, 24, 23, 25, 31, 30, 29, colored in light blue) (shown in Grow Step line 4 in Alg.~\ref{algorithm1}).
    %\item %After that, the evidence forest is updated with new trees (tree with node 4, 3, 5, 6, 7, 10, 8, 9, tree with node 15, 14, 16, 17, 18, and tree with node 26, 22, 24, 23, 25, 31, 30, 29).
    %We would start next round among new trees with roots as \texttt{"4-Conference"}, \texttt{"15-Conference"} and \texttt{"26-earn"}.
    \item Next, we repeat the above-mentioned steps until all trees in the evidence forest are connected. Finally, we construct a unclipped evidence tree whose root is \texttt{"11-defeated"} (shown in Grow Step line 7 in Alg.~\ref{algorithm1}).
    %Up till now, SGS is finished which maintains informativeness and readability of the evidence.
\end{itemize}
\end{Example}

\subsubsection{Sequential Clip Searching (SCS)}
We use SCS to remove redundant subtrees in the unclipped evidence tree to obtain a concise but still readable evidence (Fig.~\ref{fig:grow-and-clip} (d)).
\textbf{SCS} uses the hybrid scores (shown in Eq.~\ref{eq:hs}) to select the subtrees of the highest priority to prune, keeping the \emph{conciseness} and \emph{readability} of the final evidence. The process of SCS strategy is as follows and the algorithm is shown in Alg.~\ref{algorithm1} Clip Step:
\begin{itemize}
    \item Step 1 (\textbf{Select a candidate subtree to clip}): We first select a candidate subtrees $S$ in the unclipped evidence tree $\mathcal{T}^{E}$.
    Any subtree $S$ of $\mathcal{T}^{E}$ that does not appear in the evidence forest $\mathcal{F}^{E}$ is a valid candidate. This ensures that any question-relevant clue words and answer words will not be clipped, guaranteeing the informativeness of the final evidence
    (shown in line 3 of Clip Step in Alg.~\ref{algorithm1}).
    \item Step 2 (\textbf{Clip the worst candidate subtree}): If the selected candidate substree whose removal achieves the maximum hybrid score ($H_{max}$) of $\mathcal{T}^{E}$, it is pruned. Note that the entire subtree including all the children nodes and descendent nodes are deleted to guarantee the readability of the evidence. If there are more than one candidate subtrees achieving the highest hybrid score, the one whose root has a minimum weight ($W^{Attn}_{min}$) with its parent, will be selected to be pruned. This step is illustrated in line 4-7 of Clip Step in Alg.~\ref{algorithm1}.
    \item Step 3 (\textbf{Construct a clipped evidence tree}): Step 2 is repeated until $M$ times, which is a hyperparameter which is tuned by experiments. When the repeating procedure ends, a clipped evidence tree $\mathcal{T}^{E}$ is constructed.
\end{itemize}

\begin{Example}[SCS]
\label{clip}
\small
% \textbf{Sequential Clip Searching.}
% Next, we begin with LCA to make SCS based on decreasing hybrid scores and increasing attention weights to ensure conciseness and readability. 
% We put subtrees except evidence forest into the clip queue, and clip the subtree with maximum hybrid score or minimum attention weight when they have equal hybrid scores. 
Continue Example ~\ref{grow}. We drive a clipped evidence tree $\mathcal{T}^{E}$ from the result of SGS.
This clipped evidence tree is an informative-yet-concise and human-readable evidence after ranking its nodes by indexes.
% there are four candidate clipped subtrees (subtree with node 32, subtree with node 20, 19, subtree with node 22, 24, 23, and subtree with node 25) whose roots are \texttt{"32-."}, \texttt{"20-Carolina"}, \texttt{"22-24"} and \texttt{"25-to"}, respectively (shown in Clip Step line 3 in Alg.~\ref{algorithm1}).
\begin{itemize}
    \item There are four candidate clipped subtrees $\mathcal{S}$ (subtrees rooted at node 32, 20, 22, 25, respectively).
    The hybrid scores of $\mathcal{T}^{E}$ after removing each subtrees are
    % \item We first discard each subtree, and rank the remaining nodes in the unclipped evidence tree by indexes to obtain their hybrid scores, respectively. 
    % For example, after discarding the subtree with node 22, 24, 23 in the unclipped evidence tree, we obtain a clipped evidence tree with node 3, 4, 5, 6, 7, 8, 9, 10, 11, 14, 15, 16, 17, 18, 19, 20, 21, 25, 26, 29, 30, 31, 32. The corresponding evidence is \texttt{"Football Conference (AFC) champion Denver Broncos defeated Football Conference (NFC) champion Carolina Panthers to earn Super Bowl title."}, and the hybrid score of this evidence is 0.7006.
    % In this way, the hybrid scores of each clipped evidence tree are 
    0.6439, 0.6572, 0.7006 and 0.6145, respectively. Only the subtree rooted at 22 achieves the maximum score.
    % corresponding to discarding subtree with node 32, subtree with node 20, 19, subtree with node 22, 24, 23, and subtree with node 25, successively.
    \item Next, we clip the subtree $\mathcal{S}_{wor}$ rooted at 22. 
    All the descendant nodes including nodes 23, 24 are removed from $\mathcal{T}^{E}$. After that, we repeat the above steps to clip other subtrees in the same way (colored in light gray). In this example, the clip time $M$ is set to 1 according to experiments. 
    % Therefore, the final clipped evidence tree is tree with node 3, 4, 5, 6, 7, 8, 9, 10, 11, 14, 15, 16, 17, 18, 19, 20, 21, 25, 26, 29, 30, 31, 32.
    % Up till now, SCS is finished which maintains conciseness and readability of the evidence.
\end{itemize}
\end{Example}
After finishing Grow-and-Clip, we rearrange nodes in the final clipped evidence tree $\mathcal{T}^{E}$ in terms of the indexes of nodes to obtain the optimal evidence: \texttt{"Football Conference (AFC) champion Denver Broncos defeated Football Conference (NFC) champion Carolina Panthers to earn Super Bowl title."}, which is an informative-yet-concise and human-readable evidence.

\SetKwInput{GrowStep}{Grow Step}
\SetKwInput{ClipStep}{Clip Step}
\SetKwInOut{Input}{Input}
\SetKwInOut{Output}{Output}
\begin{algorithm}[t]
\DontPrintSemicolon
\caption{The Grow-and-Clip strategy}
\label{algorithm1}
\Input{Evidence Forest $\mathcal{F}^{E} = \{\mathcal{T}_1, \cdots, \mathcal{T}_N\}$,
      \newline Weighted Syntactic Parsing Tree $\mathcal{T}$,
      \newline Times of clip $M$.}
\Output{Evidence Tree $\mathcal{T}^{E}$.}
\vspace{2.5mm}
\GrowStep{}
\begin{algorithmic}[1]
\small
\STATE $N\leftarrow |\mathcal{F}^{E}|$ \tcp*[l]{$N$ is the number of trees in $\mathcal{F}^{E}$}
\WHILE{$N>1$}
% \STATE Rank the root nodes of the trees in $\mathcal{F}^{E}$ in descending order based on the attention weights $W^{attn}$
\STATE $\mathcal{T}_{opt} \leftarrow argmax_{\mathcal{T}_i\in \mathcal{F}^{E}}W^{attn}_{r}(\mathcal{F}^{E},\mathcal{T})$
\tcp*[l]{$\mathcal{T}_{opt}$ is the tree whose root has a maximum weight $W^{attn}_{r}$ with its parent in $\mathcal{T}$}
% \STATE Select $\mathcal{T}_{opt}$ with its root having maximum attention weight
% \STATE Let $r$ be the parent of the root of $\mathcal{T}_{opt}$
\STATE $\mathcal{T}_{opt}\leftarrow \text{ merge } \mathcal{T}_{opt} \text{ with parent node and sibling subtrees}$
\STATE $N\leftarrow |\mathcal{F}^{E}|$ \tcp*[l]{re-calculate the number of trees in $\mathcal{F}^{E}$}
\ENDWHILE
\STATE $\mathcal{T}^{E} \leftarrow \mathcal{T}_{opt}$ \tcp*[l]{now we get the unclipped evidence tree}
\end{algorithmic}
\vspace{2.5mm}
\ClipStep{}
\begin{algorithmic}[1]
% \STATE $j\leftarrow 1$\;
\small
\STATE $i\leftarrow M$ \tcp*[l]{how many times we can clip}
\WHILE{$i>0$}
\STATE $\mathcal{S} \leftarrow subtrees(\mathcal{T}^{E})-trees(\mathcal{F}^{E})$
\tcp*[l]{$\mathcal{S}$ doesn't include any trees in $\mathcal{F}^{E}$}
% \STATE Rank the root nodes of subtrees $\mathcal{S}$ in $\mathcal{F}^{E}$ in ascending order based on the hybrid score $H$ and in descending order based on the attention weights $W^{attn}$
% \STATE Select $\mathcal{S}_{maxlow}$ with its root having maximum hybrid scores or minimum attention weight when hybrid scores are equal 
\STATE $\mathcal{S}_{wor} \leftarrow argmax_{\mathcal{S}_i\in \mathcal{S}}(H(\mathcal{S}_i,\mathcal{T}))$
\tcp*[l]{$\mathcal{S}_{wor}$ is the subtree of maximum hybrid score $H$}
% \STATE $V \leftarrow root(\mathcal{T}^{E})$
% \STATE Stack $S \leftarrow [ V ]$ \tcp*[l]{a stack to store visited nodes}
% \WHILE{$V$ is not a leaf node}
%     \STATE $V \leftarrow \text{minAttnWeight}(\text{maxHybScore}(V.children))$
%     \STATE $S \leftarrow S + V$
% \ENDWHILE
%\STATE $S_{clip} = [\ ]$  \tcp*[l]{a stack to store clipped subtrees}
% \WHILE{$S_{clip}+ pop(S)$ doesn't include any trees in $\mathcal{F}^{E}$}
% \STATE  $S_{clip} \leftarrow  S_{clip} + \mathcal{S}_{optimal}$
% \tcp*[l]{add clipped subtree into the stack}
% \ENDWHILE
\STATE $\mathcal{T}^{E} \leftarrow \mathcal{T}^{E}-\mathcal{S}_{wor}$
% \STATE remove all nodes in $S_{clip}$ from $\mathcal{T}^{E}$
\tcp*[l]{clip the redundant evidence tree}
\STATE $i \leftarrow i-1 $
\ENDWHILE
% \STATE $\mathcal{T}^{E} \leftarrow \mathcal{T}^{E}$ 
\STATE Output $\mathcal{T}^{E}$
\tcp*[l]{now we get the clipped evidence tree}
\end{algorithmic}
\end{algorithm}

\section{Experiments}
\label{experiments}
In this section, we carry out experiments on two reading comprehension datasets SQuAD and TriviaQA.
We adopt nine fine-tuned QA models on SQuAD and another nine fine-tuned QA models on TriviaQA to predict answers. Then we take these predicted answers and also the ground-truth answers as the input answers to distill evidences, respectively.
Next, we propose a human evaluation protocol to access informativeness, conciseness and readability of the distilled evidences on these datasets.
Moreover, we use the distilled evidences as the contexts to investigate whether an off-the-shelf QA model could benefit from our distilled evidence.
In addition, we carry out ablation study to study the effect of each component of our GCED algorithm. 
We also showcase the effectiveness of distilled evidence through case studies and conduct error analysis to explore the limits of our approach.
% Our implementation code is uploaded to Github: \url{https://github.com/Yukyin/ICDE}.

% Then, we elaborate implementation details of experiments and main results of it. 

\subsection{Experimental Setup}
\subsubsection{Human evaluation}
\label{human_evaluation}
We conduct human evaluation to assess informativeness, readability and conciseness of the distilled evidences. 
Specifically, we design a scoresheet with five scales (1-5 scales, 5 is the best) to evaluate the three characteristics, as shown in Tab.~\ref{tab:humans}.
\begin{table*}[!t]
\caption{A human evaluation scoresheet for the distilled evidences based on informativeness, conciseness and readability.}
    \begin{center}
        \begin{threeparttable}
            \newcommand{\tabincell}[2]{\begin{tabular}{@{}#1@{}}#2\end{tabular}}
            \begin{tabular}{l|l}
                \toprule
                \bf Criteria &(\bf Scores) \bf Contents\\
                \midrule
                Informativeness & \tabincell{l}{(5) Extremely related to the QA pair, and the input answer can be completely inferred from the evidence.\\
                (4) Generally related to the QA pair, and the input answer can be partly inferred from the evidence.\\
                (3) Generally related to the QA pairs, but the input answer can't be inferred from the evidence.\\
                (2) Only some details between the evidence and the QA pair are identical, and the input answer can't be
                inferred from the evidence.\\
                (1) The evidence is irrelevant with the QA pairs.}\\ 
                %     &
                % \tabincell{l}{5\\4\\3\\2\\1}\\
                \midrule
                Conciseness & \tabincell{l}{(5) Extremely concise.\\
                (4) Generally concise. (1-1.5 longer than the expected evidence).\\
                (3) Containing some redundant information. (1.5-2 longer than the expected evidence)\\
                (2) Containing too much redundant information. (2-3 longer than the expected evidence)\\
                (1) The evidence is the whole document. ($>$3 longer than the expected evidence)}\\
                % & \tabincell{l}{5\\4\\3\\2\\1}\\
                \midrule
                Readability  & \tabincell{l}{(5) Extremely fluent and logical.\\
                (4) Can be understood with a few grammar mistakes (1-2).\\
                (3) Can be understood by some extent, but with many grammar mistakes ($>$2).\\
                (2) Can not be understood, but some segments are fluent.\\
                (1) Not readable.}\\ 
                % & \tabincell{l}{5\\4\\3\\2\\1}\\
                \bottomrule
            \end{tabular}
        \end{threeparttable}
    \end{center}
    \label{tab:humans}
\end{table*}

The procedure of human evaluation is as follows: We first enroll 9 graduate student volunteers (not including authors) as human raters to rate the distilled evidences according to the criterion specified in Tab.~\ref{tab:humans}. They are kindly to offer their help without being compensated in any form. 
We divide them into three groups, and raters in each group will be distributed with the same evidences. We randomly select 3,000 QA pairs per QA model per dataset for them to evaluate.  
Next, we calculate Inter-rater agreement of Krippendorff’s Alpha (IRA) shown in Tab.~\ref{tab:agreement} to evaluate the rating quality. Some controversial evidences which have low agreements ($<$0.7) are discarded. Finally we average human evaluation of each group to get the final quality score of the distilled evidences. We set the weight factor of informativeness, conciseness and readability as the same.

%inter-rater agreement of Krippendorff’s Alpha可能要写出来

\begin{table}[!t]
\caption{Inter-rater agreement of Krippendorff’s Alpha in human evaluation.}
    \begin{center}
        \begin{threeparttable}
            \begin{tabular}{p{2.7cm}|p{1.1cm}<{\centering}p{1.1cm}<{\centering}p{1.1cm}<{\centering}}
                \toprule
                \bf Criteria &\bf Group 1&\bf Group 2&\bf Group 3\\
                \midrule
                Informativeness & 0.77& 0.81& 0.76\\
                Conciseness  & 0.83 & 0.80&0.75\\
                Readability & 0.82 &0.77&0.81 \\
                \midrule
                Hybrid Score &0.81&0.79&0.78\\
                \bottomrule
            \end{tabular}
        \end{threeparttable}
    \end{center}
    \label{tab:agreement}
    \vspace{-4mm}
\end{table}

\subsubsection{QA models performance} 
We carry out experiments to observe the effect of distilled evidences for improving the performance of QA models.
we carry out experiments on GeForce RTX 3090 GPU.
For some QA models with too many parameters, we use TPU on Google Colab. 
We use \textit{Stanfordcorenlp} and \textit{nltk} to do syntactic parsing, and use \textit{Pytorch} to train QA models.
We set the maximum length of the input sequence as 384 with 64 for the question, 30 for the answer, 100 for the evidence, and 128 for the sliding windows in the context. 
Moreover, we use Adam optimizer with default parameters, initialize the learning rate and batch size to 5e-5 and 8, respectively, for 3 epochs.

\subsection{Baselines and Metrics}
\subsubsection{Baselines}
We select nine fine-tuned QA models on SQuAD, including BERT~\cite{devlin:18}, RoBERTa~\cite{RoBERTa}, SpanBERT~\cite{SpanBERT}, ALBERT~\cite{Lan:20}, XLNet~\cite{Yang:19}, ELECTRA~\cite{Kevin:20}, LUKE~\cite{Yamada:20}, T5~\cite{T5} and DeBERTa~\cite{DEBERTA}.
And select another nine fine-tuned QA models on TriviaQA, including BERT+BM25~\cite{devlin:18, The_probabilistic_relevance}, GraphRetriever~\cite{Knowledge_Guided_Text}, Longformer~\cite{Longformer}, BIGBIRD~\cite{Big_Bird}, RAG~\cite{Patrick:20}, PA+PDR~\cite{Cheng:20,Cheng:21} and Hard-EM~\cite{Min:19}, to predict answers in two situations:
\begin{itemize}
\item The input of QA models are the questions and the contexts, and the output is the answers predicted by QA models.
These predicted answers are used to distill predicted-answer-based evidences. 
\item The input of QA models is the questions and the distilled evidences based on given answers, and the output is the answers predicted by the QA models.
These predicted answers are used to evaluate the performance of QA models.
\end{itemize}

\subsubsection{Metrics}
The metrics includes human evaluation and machine evaluation shown as follows:
\begin{itemize}
\item \textbf{Human evaluation.}
We adopt human evaluation, as introduced in Sec.~\ref{human_evaluation}, to assess the quality (informativeness, conciseness, readability) of the distilled evidences.
% Inspired by model evaluation in SQuAD~\cite{squad:16}, 
\item \textbf{Machine evaluation.}
We use EM (Exact Match) and F1 (F1 score) to evaluate the performance of QA models after using the distilled evidences as the contexts. 
% We evaluate whether these ground-truth-based-evidences can be a form of ground-truth-explainable and refined contexts which can improve the performance of QA models. 
% Both metrics ignore punctuations and articles (a, an, the).
EM and F1 are same as the metrics introduced in ~\cite{squad:16,squad:18}, which measures the percentage of predictions that match any one of the ground-truth answers exactly and the average overlap between the predicted answer and the ground-truth answer, respectively. 
% We treat the predicted answers and the ground-truth answer as bags of tokens to compute F1, and we take the maximum F1 over all ground-truth answers for a question.
\end{itemize}

\subsection{Datasets}
% \textbf{Datasets.} 

We adopt two popular reading comprehension datasets, SQuAD and TriviaQA, for the evaluation. The statistics of datasets are shown in Tab.~\ref{tab:statistics}. Each dataset is divided into training set (for QA models training) and development set (for QA models evaluation).  

\textbf{SQuAD}~\cite{squad:16,squad:18} have two versions: 1.1 and 2.0, which are both derived from Wikipedia articles. SQuAD-1.1 consists of 107,785 question-answer pairs composed from 536 articles by crowdworkers. SQuAD-2.0 combines the existing questions in SQuAD-1.1 with 53,775 new unanswerable questions about the same paragraphs. The answer to every question in SQuAD-1.1 and 2.0 is a segment of text or a span from the corresponding paragraph.

\textbf{TriviaQA} \cite{triviaqa:17} contains over 650K question-answer-evidence triples. After removing the triples missing the correct answer, there remains 95K useful question-answer pairs. The dataset is constructed from Web search results and Wikipedia articles.
Hence, we have two versions of datasets: TriviaQA-Web and TriviaQA-Wiki.
%reading comprehension domain, containing 1.8 evidence documents per example on average. Specifically, we use their unfiltered set.

%看一下TriviaQA的答案到底需要涉及多个句子还是需要推理，怎么样推理，是每个句子之间有联系吗，怎样的联系，多个句子都包含答案吗？所以这就是结果不好的原因。需要说明下如果答案需要多个句子推理（不需要知识库），怎么搞？把第一个answer select extractor模块改成attentionbilstm什么的抽取候选句子，之后对每个句子用后续方法。这样可以解决多步推理问题。trivaqa效果不好应该是冗余信息太多了。

\begin{table}[!t]
\caption{Statistics of two datasets, SQuAD and TriviaQA. 
% We apply SQuAD-1.1,  SQuAD-2.0, Web-based TriviaQA and Wikipedia-based TriviaQA to carry out experiments.
}
    \begin{center}
        \begin{threeparttable}
            \begin{tabular}{p{2.4cm}|p{1.3cm}<{\centering}p{1.3cm}<{\centering}}
                \toprule
                \bf Dataset &\bf Train&\bf Dev\\
                \midrule
                SQuAD-1.1 & 87599 & 10570\\
                SQuAD-2.0  & 130319 & 6078\\
                TriviaQA-Wikipedia & 110647 &14229 \\
                TriviaQA-Web &100000  & 68621\\
                \bottomrule
            \end{tabular}
        \end{threeparttable}
    \end{center}
    \label{tab:statistics}
    \vspace{-4mm}
\end{table}

% \textbf{Models Variants}
% \begin{itemize}
%   \item \textbf{BERT\citep{devlin:18}} is the powerful pre-trained large-BERT which has 24 transformer blocks, 1024 dimension of hidden state, 16 head per layer of multi-head attention and 110M parameters. We separate questions and contexts by \verb"[SEP]" to join them to predict answers.
%   \item \textbf{AS-simp} is to substitute answer-oriented sentences for contexts in the aforementioned BERT to join with questions to predict answers.
%   \item \textbf{AS-aug} is joining questions with the combination of contexts and answer-oriented sentences in the aforementioned BERT to predict answers.
%   \item \textbf{GCED-simp} is to substitute our distilled evidences for contexts to join with questions in the aforementioned BERT to predict answers.
%   \item \textbf{GCED-aug} is joining questions with the combination of contexts and our distilled evidences in the aforementioned BERT to predict answers. The details are shown in Fig.~\ref{fig:model}.
% \end{itemize}

% \begin{figure}
%   \centering
%   \includegraphics[width=\linewidth]{model.png}
%   \caption{GCED-augmented question answering model with a combination of questions, contexts and evidence to predict answers.}
%   \label{fig:model}
% \end{figure}

\subsection{Main results}
\label{Mainresults}
Next, we present the main results of our experimental study.
\subsubsection{Results of human evaluation}
\label{Results of human evaluation}
The results of human evaluation for distilled evidences are shown in Tab.~\ref{tab:squad_human} (on SQuAD) and Tab.~\ref{tab:tqa_human} (on TriviaQA). 
We make human evaluation for both predicted-answer-based evidences (row 1-9 in Tab.~\ref{tab:squad_human} and Tab.~\ref{tab:tqa_human}) and ground-truth-answer-based evidences (the last row in Tab.~\ref{tab:squad_human} and Tab.~\ref{tab:tqa_human}). 

The human evaluation reveals that for both predicted-answer-based evidences and ground-truth-based evidences, the quality scores (informativeness, conciseness, readability as well as the hybrid scores) of distilled evidences are consistently larger than 0.75 across all baseline QA models and datasets. It suggests that the quality of automatically distilled evidences is satisfying, which is independent on different QA models.
We also find that on average 78.5\% and 87.2\% words have been reduced in the distilled evidences on SQuAD and TriviaQA datasets, respectively. This further justifies the conciseness of the evidences. 
% Moreover, we make human evaluation for the distilled evidences which are generated based on the ground-truth answers besides based on the predicted answers.
We highlight that there is no significant difference between human evaluation for predicted-answer-based evidences and ground-truth-based evidences (the p-value is $>0.5$). 
The reason is that the distilled evidences aim at explaining/supporting the input answers whatever the input answers are predicted by QA models or labeled as ground-truth answers by crowdworkers.
The correctness of the answers do not affect the quality of the distilled evidences, which is friendly for humans to understand the source of the input answers.

\begin{table}[!t]
\caption{Human evaluation for predicted-answer-based evidences and ground-truth-answer-based evidences on SQuAD-1.1 and SQuAD-2.0. The predicted-answer-based evidences are distilled from the answers predicted by the QA models listed in the first column. 
}
    \begin{center}
        \begin{threeparttable}
            \setlength{\tabcolsep}{1mm}{
                \begin{tabular}{l|llll|llll}
                    \toprule \bf Datasets& \multicolumn{4}{c|}{\bf SQuAD-1.1}& \multicolumn{4}{c}{\bf SQuAD-2.0}\\
                    \midrule
                    \bf Source & \bf I& \bf C& \bf R& \bf H& \bf I& \bf C& \bf R& \bf H\\
                    \midrule
                    BERT-large &0.87&0.82&0.83&\textbf{0.84}&0.88&0.81&0.87&\textbf{0.85}\\
                    RoBERTa-500K&0.86&0.89&0.84&\textbf{0.86}&0.85&0.90&0.88&\textbf{0.88}\\
                    SpanBERT&0.88&0.86&0.87&\textbf{0.87}&0.85&0.82&0.84&\textbf{0.84}\\
                    ALBERT&0.88&0.83&0.86&\textbf{0.86}&0.89&0.84&0.83&\textbf{0.86}\\                  XLNet-large&0.90&0.87&0.88&\textbf{0.88}&0.90&0.89&0.89&\textbf{0.89}\\
                    ELECTRA-1.75M&0.91&0.87&0.85&\textbf{0.88}&0.91&0.85&0.88&\textbf{0.88}\\
                    LUKE&0.86&0.83&0.85&\textbf{0.85}&0.90&0.86&0.89&\textbf{0.88}\\
                    T5&0.89&0.87&0.86&\textbf{0.87}&0.92&0.89&0.88&\textbf{0.90}\\
                    DeBERTa-large&0.90&0.86&0.82&\textbf{0.86}&0.89&0.88&0.90&\textbf{0.89}\\
                    % \midrule                    (Average)&0.88&0.86&0.86&0.87&0.89&0.86&0.87&0.87\\
                    \midrule                                      Ground-truth&0.90&0.89&0.88&\textbf{0.89}&0.91&0.90&0.89&\textbf{0.90}\\
                    \bottomrule
            \end{tabular}}
        \end{threeparttable}
    \end{center}
    \label{tab:squad_human}
    \vspace{-4mm}
\end{table}

\begin{table}[!t]
\caption{
Human evaluation for predicted-answer-based evidences and ground-truth-based evidences on TriviaQA datasets. The predicted-answer-based evidences are distilled based on the answers predicted by the following QA models. }
    \begin{center}
        \begin{threeparttable}
            \setlength{\tabcolsep}{1mm}{
                \begin{tabular}{l|llll|llll}
                    \toprule
                    \bf Datasets& \multicolumn{4}{c|}{\bf TriviaQA-Web}& \multicolumn{4}{c}{\bf TriviaQA-Wiki}\\
                    \midrule
                    \bf Source & \bf I& \bf C& \bf R& \bf H& \bf I& \bf C& \bf R& \bf H\\
                    \midrule
                    BERT+BM25 &0.82&0.81&0.80&\textbf{0.81}&0.83&0.82&0.81&\textbf{0.82}\\
                    GraphRetriever&0.81&0.82&0.77&\textbf{0.80}&0.77&0.81&0.76&\textbf{0.78}\\
                    RoBERTa-base&0.83&0.85&0.80&\textbf{0.83}&0.82&0.78&0.80&\textbf{0.80}\\
                    Longformer-base&0.81&0.77&0.79&\textbf{0.79}&0.79&0.76&0.76&\textbf{0.77}\\
                    Bigbird-itc&0.77&0.80&0.77&\textbf{0.78}&0.83&0.76&0.78&\textbf{0.79}\\
                    ELECTRA-base&0.85&0.86&0.83&\textbf{0.84}&0.81&0.82&0.80&\textbf{0.81}\\
                    RAG-Sequence&0.83&0.78&0.79&\textbf{0.80}&0.80&0.84&0.82&\textbf{0.82}\\
                    PA+PDR&0.80&0.84&0.82&\textbf{0.82}&0.80&0.82&0.80&\textbf{0.80}\\
                    Hard-EM&0.85&0.86&0.81&\textbf{0.83}&0.83&0.80&0.81&\textbf{0.81}\\
                    \midrule
                    % (Average)&&&&&&&&\\
                    % \midrule
                    Ground-truth&0.85&0.86&0.84&\textbf{0.85}&0.83&0.84&0.82&\textbf{0.83}\\
                    \bottomrule
            \end{tabular}}
        \end{threeparttable}
    \end{center}
    \label{tab:tqa_human}
    \vspace{-4mm}
\end{table}

\subsubsection{Performance gain of QA models augmented by ground-truth-answer-based evidences}
\label{performance_gain}
In an ideal setting or some applications (such as searching engine), we have ground-truth answers, which are ideal sources to distill evidences. These evidences could also be seen as concise-yet-informative contexts for a QA model to find the correct answers.
If the evidence is more concise than the original context, and it contains essential information to answer a question, then using the evidence as the input of a QA model (instead of the raw context) shall improve its performance. 
Hence, we propose an experiment to test the performance gain of QA modes which are augmented by evidence distilled by our solution from the ground-truth-answers. 

% We observed the performance improvement, justifying our hypothesis about the evidences.
% , if our solution is able to distill effective evidences from the ground-truth answer. In contrast, a bad evidence serving as the context could significantly hurt the QA model. 

The results are shown in Tab.~\ref{tab:squad_results} (SQuAD) and Tab.~\ref{tab:tqa_results} (TriviaQA).
Results of baselines are from their published papers except those with asterisks in the upper right corner, which are missing in the corresponding papers. We retrain these baselines with asterisks based on the same settings on our own machine and report their results.
We find that it performs consistently better for all QA models when the contexts are replaced with the distilled evidences on all tested datasets, which supports our conjecture.

\subsubsection{Performance degradation of QA models augmented by predicted-answer-based evidences}
\label{performance_degradation}
In a more realistic setting, we have no ground-truth answers, where we need to predict the answer by a vanilla QA model. However, the evidence distilled from a wrong predicted answer might be noisy for the question answering.
Using such evidences as the QA contexts is likely to produce wrong answers. Hence, QA models augmented by predicted-answer-based evidences are expected to degrade in their performance. Hopefully, the performance degradation is minor or acceptable if our evidence distillation solution is effective to finding effective information from the answer. This inspires our interest to test the performance degradation of QA models which are augmented by the evidences distilled by our solution from the predicted answers. 
% However, \emph{it is still interesting to see whether the performance degradation is acceptable.}
% not effective for the question answering, the performance of evidence distillation from predicted answer is must inferior to that of the QA model.
% pit is necessary to verify how the QA models degrade when it utilizes predicted answers instead of the ground-truth ones to distill evidences.

\begin{figure*}[!t]
	\centering
	\includegraphics[width=\linewidth]{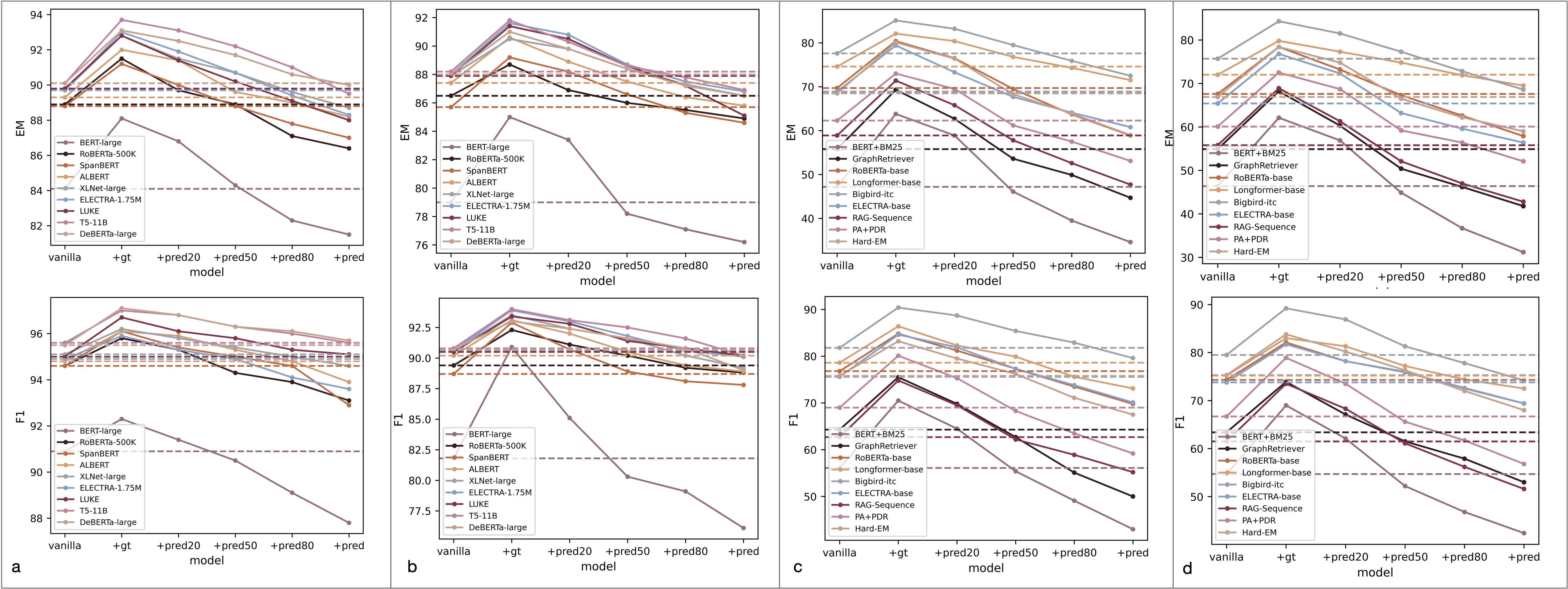}
	\caption{The performance degradation of QA models augmented by predicted-answer-based evidences (a: SQuAD-1.1, b: SQuAD-2.0, c: TriviaQA-Web and d: TriviaQA-Wiki). gt: use ground-truth answers to distill evidences; pred: use predicted answers to distill evidences; predx: substitute x\% ground-truth answers with predicted answers to distill evidences.}
	\label{fig:evidence-bias}
\end{figure*}

We randomly substitute $\delta$ percentage of ground-truth answers with predicted answers with $\delta$ ranging in {0.2, 0.5, 0.8, 1}, respectively, to distill evidences.
It means some distilled evidences are based on the ground-truth answers and others are based on the predicted answers. 
We then substitute the contexts with these evidences to retrain QA models.
The results are shown in Fig.~\ref{fig:evidence-bias}.
% In order to avoid exposure bias, the substitution acts on both training set and dev set.
% We analyze evidence bias on all baselines and datasets.
% We show the gap between ground-truth-based evidence and prediction-based one in Fig.~\ref{fig:evidence-bias}.
%, including results on SQuAD-1.1, SQuAD-2.0, TriviaQA-Web and TriviaQA-Wiki datasets with EM and F1.	
It's obvious to find that the performance of QA models fed with the predicted-answer-based evidences as the contexts indeed degrade compared with that of the vanilla QA models, which confirms our conjecture. However, for many QA models, even all the evidences come from predicted answer, only 2-3\% performance drop can be observed on SQuAD datasets. This result suggests our solution has minor side effects but additionally generates answering evidence, which are demanded in many real applications.
More performance degradation can be observed on TriviaQA dataset. Because this dataset is more open and in general is more challenging, and most State-Of-The-Art QA models have less than 80\% performance (EM \& F1). 

This could be an interesting research topic in the future work. Whatever the answer is correct or not, the corresponding evidence provides an informative-yet-concise summarization of the context that explains how this answer was predicted. In this way, a user can quickly know the information source of this answer, leading to an explainable and reliable QA system. For example, given a question \texttt{"Where was Albert Einstein born?"}, if a QA system gives the answer \texttt{"Berlin"} and the supporting evidence \texttt{"Albert Einstein moved to Berlin 50 years old."}, the user will find this QA system is not reliable and he will not trust this answer.

% Moreover, 
% the gap is positively correlated with the performance of vanilla QA models.
% That is to say, the more accurate the QA models, the much closer the predicted answers to the ground-truth answers. 
% For example, when $\delta=20\%$, almost each evidence-based model outperforms vanilla one except the worst BERT.
% As for the best ALBERT, XLNet, ELECTRA, $\delta=50\%$ can still achieve a high accuracy.
% When the $\delta=80\%$, although all evidence-based models are inferior to vanilla ones, the performance degradation is minor and acceptable.

% Therefore, the quality of the evidence distilled in terms of predicted answer is comparable with that from ground-truth answers. This suggests that our solution is also suitable for the more realistic setting where the answer predicted by a QA model is used for the evidence distillation. In other words, GCED can simultaneously generate answers and their supporting evidences. We also highlight that the distilled evidences is a more informative-yet-concise representation of the original contexts.

%speed and answer interpretability without sacrificing accuracy.

\begin{table}[!t]
\caption{Comparisons of nine baselines and their evidence-augmented variants on SQuAD-1.1 and SQuAD-2.0 datasets. The evidences are distilled from the ground-truth answers. 
}
    \begin{center}
        \begin{threeparttable}
            \setlength{\tabcolsep}{1mm}{
                \begin{tabular}{p{2.4cm}p{1.2cm}<{\centering}p{1.2cm}<{\centering}p{1.2cm}<{\centering}p{1.2cm}<{\centering}}
                    \toprule                    \bf Datasets& \multicolumn{2}{c}{\bf SQuAD-1.1}& \multicolumn{2}{c}{\bf SQuAD-2.0}\\
                    \midrule
                    \bf Models & \bf EM& \bf F1& \bf EM& \bf F1\\
                    \midrule
                    BERT-large &84.1&90.9&79.0 &	81.8 	\\
                    \qquad \bf+GCED &\bf 88.1&\bf92.3&\bf85.0&\bf90.9\\
                    \midrule
                    RoBERTa-500K&88.9&94.6&86.5&89.4\\
                    \qquad \bf+GCED &\bf91.5&\bf95.8&\bf88.7&\bf92.3\\
                    \midrule
                    SpanBERT&88.8&94.6&85.7&88.7\\
                    \qquad \bf+GCED &\bf 91.2&\bf 96.1&\bf89.2&\bf92.9\\
                    \midrule
                    ALBERT&89.3&94.8&87.4&90.2\\
                    \qquad \bf+GCED &\bf92.0&\bf96.1&\bf90.6&\bf93.1\\
                    \midrule
                    XLNet-large&89.7&95.1&87.9&90.6\\
                    \qquad \bf+GCED &\bf92.8&\bf96.2&\bf90.5&\bf93.5\\
                    \midrule
                    ELECTRA-1.75M&89.7&94.9&88.0&90.6\\
                    \qquad \bf+GCED &\bf93.0&\bf95.9&\bf91.6&\bf93.9\\
                    \midrule
                    LUKE&89.8&95&87.9*&90.5*\\
                    \qquad \bf+GCED &\bf92.8&\bf96.7&\bf91.4&\bf93.4\\
                    \midrule
                    T5&90.1&95.6&88.2*&90.8*\\
                    \qquad \bf+GCED &\bf93.7&\bf97.0&\bf91.8&\bf94.0\\
                    \midrule
                    DeBERTa-large&90.1&95.5&88.0&90.7\\
                    \qquad \bf+GCED &\bf93.1&\bf97.1&\bf91.0&\bf93.0\\
                    \midrule
                    \bf+GCED (average) &\bf$\uparrow$\bf3.5\%&\bf$\uparrow$\bf1.5\%&\bf$\uparrow$\bf4.1\%&\bf$\uparrow$\bf4.2\%\\
                    \bottomrule
            \end{tabular}}
        \end{threeparttable}
    \end{center}
    \label{tab:squad_results}
    \vspace{-4mm}
\end{table}

\begin{table}[!t]
\caption{Comparisons of nine baselines and their evidence-augmented variants on TriviaQA datasets. The evidences are distilled from the ground-truth answers.}
    \begin{center}
        \begin{threeparttable}
            \setlength{\tabcolsep}{1mm}{
                \begin{tabular}{p{2.4cm}p{1.2cm}<{\centering}p{1.2cm}<{\centering}p{1.2cm}<{\centering}p{1.2cm}<{\centering}}
                    \toprule
                    \bf Datasets&\multicolumn{2}{c}{\bf TriviaQA-Web}& \multicolumn{2}{c}{\bf TriviaQA-Wiki}\\
                    \midrule
                    \bf Models & \bf EM& \bf F1& \bf EM& \bf F1\\
                    \midrule
                    BERT+BM25 &47.2&56.1*&46.4*&54.7*\\
                    \qquad \bf+GCED &\bf63.8&\bf70.5&\bf62.1&\bf69.0\\
                    \midrule
                    GraphRetriever&55.8&64.3*&54.9*&63.4*\\
                    \qquad \bf+GCED &\bf69.3&\bf75.5&\bf68.2&\bf73.9\\
                    \midrule
                    RoBERTa-base&69.7*&76.8*&67.6*&74.3\\
                    \qquad \bf+GCED &\bf80.4&\bf84.8&\bf78.4&\bf82.1\\
                    \midrule                    Longformer-base&74.6*&78.6*&72.0*&75.2\\
                    \qquad \bf+GCED &\bf82.1&\bf86.4&\bf79.8&\bf83.0\\
                    \midrule
                    Bigbird-itc&77.6*&81.8*&75.7*&79.5\\
                    \qquad \bf+GCED &\bf85.1 &\bf90.4&\bf84.3&\bf89.2\\
                    \midrule
                    ELECTRA-base&68.9*&75.6*&65.4&73.8*\\
                    \qquad \bf+GCED &\bf79.4&\bf84.6&\bf76.8&\bf81.7\\
                    \midrule
                    RAG-Sequence&58.9*&62.7*&55.8&61.5*\\
                    \qquad \bf+GCED &\bf71.4&\bf74.8&\bf68.9&\bf73.5\\
                    \midrule
                    PA+PDR&62.3*&69.0*&60.1&66.7*\\
                    \qquad \bf+GCED &\bf73.0&\bf80.1&\bf72.5&\bf78.9\\
                    \midrule
                    Hard-EM&68.5*&75.8*&66.9&75.3*\\
                    \qquad \bf+GCED &\bf80.1&\bf83.2&\bf78.4&\bf83.8\\
                    \midrule
                     \bf+GCED (average) &\bf$\uparrow$\bf18.2\%&\bf$\uparrow$\bf14.6\%&\bf$\uparrow$\bf19.3\%&\bf$\uparrow$\bf15.0\%\\
                    \bottomrule
            \end{tabular}}
        \end{threeparttable}
    \end{center}
    \label{tab:tqa_results}
\end{table}

% \begin{table}[!t]
% \caption{Human evaluation of distilled evidence on SQuAD-1.1, SQuAD-2.0, TriviaQA-Web and TriviaQA-Wiki.}
%     \begin{center}
%         \begin{threeparttable}
%             \setlength{\tabcolsep}{1mm}{
%                 \begin{tabular}{p{2.2cm}|p{2.5cm}<{\centering}}
%                     \toprule
%                     \bf Datasets & \bf Human scores\\
%                     \midrule
%                     SQuAD-1.1 &0.88\\
%                     SQuAD-2.0 &0.91\\
%                     TriviaQA-Web&0.84\\
%                     TriviaQA-Wiki&0.82\\
%                     \bottomrule
%             \end{tabular}}
%         \end{threeparttable}
%     \end{center}
%     \label{tab:humanscores}
% \end{table}

\subsection{Ablation Study}
In this part, we conduct ablation study to further analyze each component of GCED. Specifically, we use BERT which is augmented by the ground-truth-based evidences on SQuAD-2.0 to make analysis. The results are shown in Tab.~\ref{tab:ablationstudy}.

\subsubsection{Ablation Study on human evaluation}
First, when we remove QWS and the informativeness score (\textbf{w/o I}), respectively, we observe that the human informativeness scores (\textbf{I}) decrease. 
It indicates that question-relevant clue words in QWS and $PLM$ in building informativeness scores both extract useful information which can explain/support the input answers.
Next, we can find that when we remove ASE (\textbf{w/o ASE}), the clip step (\textbf{w/o Clip}) from the Grow-and-Clip strategy, and the conciseness score (\textbf{w/o C}), respectively, the human conciseness scores (\textbf{C}) decrease. 
It indicates that each of these three modules can filter out much redundant information which maintains the conciseness of the distilled evidences.
After that, when we remove grow step (\textbf{w/o Grow}) from the Grow-and-Clip strategy and the readability score (\textbf{w/o C}) from the hybrid score, respectively, the human readability scores (\textbf{C}) decrease.
It indicates that both of the grow step and the readability scores can guarantee the readability of the distilled evidences.
Therefore, human evaluation of each component in the proposed GCED demonstrates that each component and the organic combination of all components in GCED both have significant effects for distilling informative-yet-concise and human-readable evidences.

\subsubsection{Ablation Study on QA models which are augmented by ground-truth-based evidences}
Next, we make ablation Study on QA models which are augmented by ground-truth-based evidences.
We can find that ASE affects most on QA model's performance. When it's removed (\textbf{w/o ASE}), the performance decreases most. 
% by 18.1\% in EM and 16.2\% in F1 compared with \textbf{BERT+GCED}. 
This results demonstrate that the answer-oriented sentence(s) contain suggestive information in predicting correct answers if we use ground-truth-answer-based evidences as the contexts to train QA models. 
Next, we remove QWS (\textbf{w/o QWS}), and the performance also decreases a lot. 
% by 21.1\% in EM and 18.8\% in F1 compared with \textbf{BERT+GCED}).
The results demonstrate that question-relevant clue words have a significant semantic relationship with the ground-truth answers.
After that, we remove Grow step (\textbf{w/o Grow}) and Clip step (\textbf{w/o Clip}) in the Grow-and-Clip strategy.
% which means directly taking words in the evidence forest as the final evidence.
% , and GC to LCA (Least Common Ancestor) which represents abandoning dependency relationship between LCA and its other children besides DF (dependency forest). 
The degrading performance 
% results 
% decrease by 13.0\% in EM and 12.8\% in F1 as well as 5.6\% in EM and 5.3\% in F1 compared with \textbf{BERT+GCED}, respectively, 
demonstrates that both of Grow step and Clip step have critical effects in predicting correct answers.
% \textbf{GC} is critical to generate a more informative, consice and readable evidence.
% and outperforms LCA 
% , and GC to LCA decreases by 4.64\% in F1 and 5.63\% in EM), 
% because LCA primarily pays attention to extract scattered continuous fragment locally ignorant of some dependencies, 
% because \textbf{GC} centers globally on the whole answer-oriented sentence(s) with connecting more continuous segments to satisfy both models and humans.
Furthermore, we assess the individual effect of the informativeness scores (\textbf{w/o I}), the conciseness scores (\textbf{w/o C}), and the readability scores (\textbf{w/o R}). 
The results demonstrate all three criteria have positive effect in predicting correct answers if these evidences are distilled based on the ground-truth answers.
% The gap between \textbf{w/o C} and \textbf{BERT+GCED} is larger than that between \textbf{w/o R} and \textbf{BERT+GCED}, because conciseness represents discarding redundant information so as to retain more useful information which is helpful for predicting correct answers in the ideal situation
% while readability aims mostly at generating a more human-readable evidence. 
% And informativeness also has a positive effect in predicting correct answers in the ideal situation.

% Besides, we also change GC to G (w/o C), and the results verify the importance of Clip strategy which focuses on removing abundant information.

% Moreover, it's worth noting that if we eliminate QWS, DF and GC, keeping the defensive modifier instead (w/o QDG (+) and w/o QDG (-)), the performance is stunning in F1 and EM but suffer in human evaluation compared with our proposed GCED-aug. w/o QDG (+) represents expanding the border of answers until obtaining wrong predictions. And w/o QDG (-) means deleting random words in answer-oriented sentences except for answers until making mistakes in prediction. In fact, the above two both maintain \verb"prep" (\emph{of}, \emph{by}) or \verb"art" (\emph{a}, \emph{the}) joint with answers, which actually cheat models, because they are almost indeed answers without any information in contexts or questions.
In brief, the proposed GCED is a powerful algorithm in distilling informative-yet-concise and human-readable evidences, which can explain/support the input answers.
And if the evidences are distilled based on the ground-truth answers, these evidences are an informative-yet-concise form of the contexts, which can improve the performance of QA models besides explaining/supporting the input answers.
% the most important components are \textbf{Answer-oriented Sentence Extractor (ASE)}, \textbf{Query-based Words Selector (QWS)} and \textbf{Grow-and-Clip strategy}. Moreover, \textbf{Concise scores (C)} are more vital than \textbf{Informativess scores (I)} and \textbf{Readability scores (R)} in \textbf{Hybrid Scores (H)}.
% It strongly demonstrates 
% the organic combination of all components constructs a powerful evidence discovery technique GCED, which can 

\begin{table}[!t]
\caption{The effect of each component of GCED. The evidences are distilled from the ground-truth answers with BERT on SQuAD-2.0.}
    \begin{center}
        \begin{threeparttable}
            \setlength{\tabcolsep}{1mm}{
                \begin{tabular}{l|llllll}
                    \toprule
                    \bf Sources & \bf I& \bf C& \bf R& \bf H &\bf EM& \bf F1\\
                    \midrule
                    w/o ASE &0.85&0.65&0.80&0.77 &72.0 &78.2\\%-c
                    w/o QWS&0.67&0.79&0.77& 0.74&70.2 &76.5\\%-i
                    w/o Grow&0.82&0.80&0.67& 0.76&75.2	&80.6\\%-r
                    w/o Clip&0.81&0.70&0.81& 0.77&80.5 &86.3 \\%-c
                    \midrule				
                    w/o I &0.73&0.78&0.80& 0.77& 80.2&87.0\\%-i
                    w/o C &0.80&0.72&0.76& 0.76&79.3 & 86.9\\%-c
                    w/o R &0.81&0.83&0.75& 0.80& 82.1&88.4\\%-r
                    \midrule
                    % BERT &79.0&81.8\\
                    \bf BERT+GCED& \bf0.86& \bf0.83& \bf0.82& \bf0.84&\bf 85.0&\bf 90.9\\
                    \bottomrule
            \end{tabular}}
        \end{threeparttable}
    \end{center}
    \label{tab:ablationstudy}
\end{table}

%\subsection{Parameters sensitivity}%不一定要

\subsection{Case Study}
\label{case_study}
\begin{figure}[!t]
    \centering
    \includegraphics[width=0.95\linewidth]{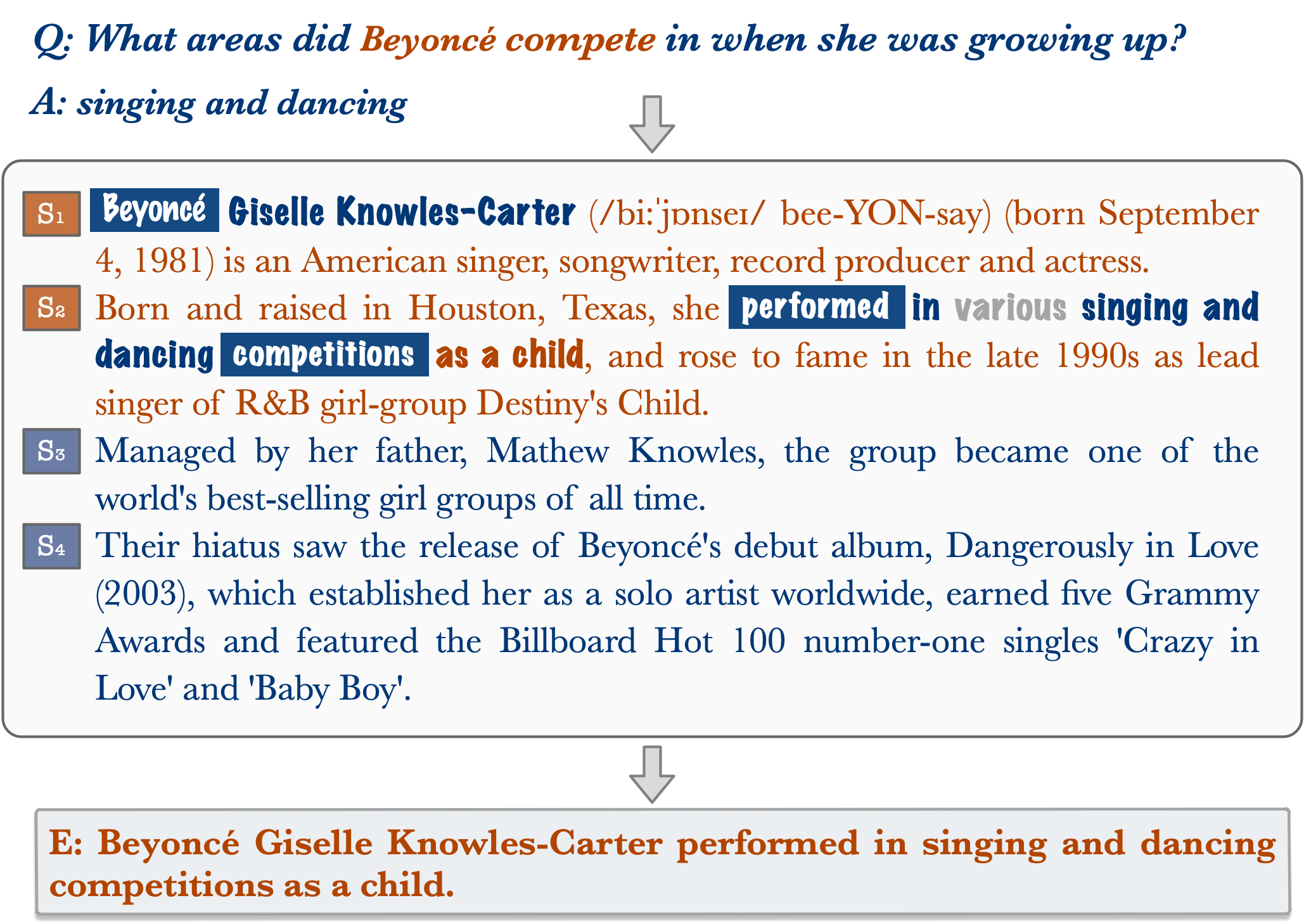}
    \caption{A good evidence distilled by GCED. Given a QA pair, $S_1$ and $S_2$ are selected as the answer-oriented sentences by ASE. 
    Question-relevant clue words (the highlighted words) are used to construct the evidence forest (bold and colored in dark blue).
    We use Grow-and-Clip strategy to grow words (bold and colored in brown) and clip words (colored in grey), and connect them to the optimal evidence.}
    \label{fig:casestudy}
    \vspace{-4mm}
\end{figure}

In this part, we display a good evidence distilled by the proposed GCED, as shown in Fig.~\ref{fig:casestudy}. 
Given a QA pair and the context which contains four sentences, $S_1$ and $S_2$ are selected as the answer-oriented sentences by ASE at first. 
Second, we select \texttt{"Beyoncé"}, \texttt{"performed"}, and \texttt{"competitions"} in the answer-oriented sentences as the question-relevant clue words by QWS.
Third, we adopt $S_1$ and $S_2$ to construct a weighted syntactic parsing tree by WSPTC with calculating attention weights for all edges.
Fourth, we obtain an evidence forest including an answer tree and two evidence trees.
The answer tree contains three nodes \texttt{"singing"}, \texttt{"and"}, \texttt{"dancing"}.
One evidence tree contains five nodes \texttt{"Beyoncé"}, \texttt{"Giselle"}, \texttt{"Knowles"}, \texttt{"-"}, and \texttt{"Carter"}.
The other evidence tree contains seven nodes \texttt{"performed"}, \texttt{"in"}, \texttt{"various"}, \texttt{"singing"}, \texttt{"and"}, \texttt{"dancing"}, and \texttt{"competitions"}.
Fifth, we adopt OEC with the Grow-and-Clip strategy to grow three nodes including \texttt{"as"}, \texttt{"a"}, and \texttt{"child"} to achieve informativeness, and clip one node including \texttt{"various"} to achieve conciseness.
Due to space limitation, we haven't display the nodes which have been grown first and then clipped, such as \texttt{"born"}, \texttt{"and"}, \texttt{"raised"}, \texttt{"in"}, \texttt{"Houston"}, \texttt{","}, \texttt{"Texas"}, etc.
These nodes mainly aim at guarantee readability during the process of Grow-and-Clip.
% including \texttt{"("}, \texttt{"/"}, \texttt{"bi"}, \texttt{":"}, \texttt{"jonsei"}, \texttt{"/"}, \texttt{"bee"}, \texttt{"-"}, \texttt{"YON"}, \texttt{"-"}, \texttt{"say"}, \texttt{")"}, \texttt{"("}, \texttt{"born"}, \texttt{"September"},  \texttt{"4"}, \texttt{"1981"}, \texttt{")"}, \texttt{"is"}, \texttt{"an"}, \texttt{"American"}, \texttt{"singer"}, \texttt{","}, \texttt{"songwriter"}, \texttt{","}, \texttt{"record"}, \texttt{"producer"}, \texttt{"and"}, \texttt{"actress"}, \texttt{","}, \texttt{"born"}, \texttt{"and"}, \texttt{"raised"}, \texttt{"in"}, \texttt{"Houston"}, \texttt{","}, \texttt{"Texas"}, and \texttt{","} to construct an unclipped evidence tree to guarantee informativeness and readable of the evidence.
% And clip 39 nodes, including \texttt{"("}, \texttt{"/"}, \texttt{"bi"}, \texttt{":"}, \texttt{"jonsei"}, \texttt{"/"}, \texttt{"bee"}, \texttt{"-"}, \texttt{"YON"}, \texttt{"-"}, \texttt{"say"}, \texttt{")"}, \texttt{"("}, \texttt{"born"}, \texttt{"September"},  \texttt{"4"}, \texttt{"1981"}, \texttt{")"}, \texttt{"is"}, \texttt{"an"}, \texttt{"American"}, \texttt{"singer"}, \texttt{","}, \texttt{"songwriter"}, \texttt{","}, \texttt{"record"}, \texttt{"producer"}, \texttt{"and"}, \texttt{"actress"}, \texttt{","}, \texttt{"born"}, \texttt{"and"}, \texttt{"raised"}, \texttt{"in"}, \texttt{"Houston"}, \texttt{","}, \texttt{"Texas"}, \texttt{","}, and \texttt{"various"} to construct a clipped evidence tree to guarantee conciseness and readable of the evidence.
Finally, we rank the remaining nodes by their indexes to distill the final optimal evidence, that is \texttt{"Beyoncé Giselle Knowles-Carter performed in singing and dancing competitions as a child"}.

We are satisfied to find that this evidence is informative (contain useful information), concise (the length is short enough) and readable (without grammar mistakes or fuzzy logic), demonstrating the proposed GCED can catch important information to explain/support the input answer, filtering out noises and is user-friendly.

\subsection{Error analysis}
\label{error_analysis}
% 可以加个图
However, there are some unsatisfying evidences distilled by GCED.
% The primary reasons mainly exist in lack of world knowledge/commonsense and reasoning.
For example, 
% given a question \texttt{"What was the first name of Russian monk Rasputin, who befriended Nicholas II and his family? }, the answer predicted by a QA model is \texttt{"Rasputin"}. The evidence distilled by GCED is \texttt{'"mad monk" Rasputin from the court of Tsar Nicholas in Russia, Nicholas and the royal doctor are both skeptical of Rasputin's abilities.'}.
% However, it doesn't explain/support the predicted answer \texttt{"Rasputin"} for the question, and only distill some related information to the QA pair.
% The possible reason is that our algorithm lacks world knowledge/commonsense in some important phrases, such as \texttt{"first name"}.
% % We generate an informative-yet-concise evidence of this QA pair by human annotators, that is 
% % Father Grigory Rasputin, a destitute monk telling him that the Tsar needed him. 
given a question \texttt{"In the Bible, who was the mother of Solomon?"}, the the answer predicted by a QA model is \texttt{"bathsheba"}. The evidence distilled by GCED is \texttt{"Solomon had brothers through Bathsheba, Nathan, Shammua, and Shobab, brothers through as many mothers."}.
However, it's difficult 
% also doesn't explain/support the predicted answer \texttt{"bathsheba"} for the question, and 
to understand by humans due to its bad readability .
Because GCED doesn't have knowledge to know the relationship among \texttt{child}, \texttt{David}, and \texttt{wife}, so it can't distill a more human-readable informative-yet-concise evidence \texttt{"Solomon was the child to David and his wife Bathsheba."}.
Moreover, some contexts are comparatively long, containing many sentences.
The structure of most sentences are complicated with nested clauses and hidden subjects, making it difficult to extract more suggestive information and filter out redundant noises to distill informative-yet-concise evidences.
% Moreover, some sentences do not have real subjects. The sentence which contains the answer will use \texttt{"pron"} to represent the real subject. The distance between \texttt{"pron"} and the real subject is too far, so it's difficult for our algorithm to extract the real subject. 
% For example, for the given QA pair \emph{``How many men were in Robert's army? 30,000"}, the answer-oriented sentence \emph{``After ..., in 1081 he led an army of 30,000 men in 300 ships..."} misses real subject, and our distilled evidence \emph{``he led an army of 30,000 men"} also ignore it. 

Therefore, we will optimize our proposed GCED through adding world knowledge and commonsense and improving the understanding on too complicated sentences in the future work.

\section{Related work}
\textbf{Evidences extraction based on sentence-level for answer explanation}.
A succession of research has been conducted to improve the interpretability of QA. For the extractive and abstractive QA, such as SQuAD \cite{squad:16,squad:18} and TriviaQA \cite{triviaqa:17}, most answers and their supporting evidences can be directly extracted from the given contexts based on semantic and syntactic knowledge. The process of evidences extraction can be summarized as extracting only a few sentences in the documents to explain/support the answer. 
Previous studies mainly extract evidences on sentence level. For example, Min et al. \cite{min:18} extract the minimal context based on a sentence selector and a QA model, giving the QA model a reduced set of sentences with high selection scores in order to explain/support answers.
Yin et al. \cite{Yin:16} study interdependent sentence pair representations with three attention schemes, integrating mutual influence between sentences into CNNs to explain/support answers on WikiQA \cite{yi:15}. 
Choi et al. \cite{choi:17} combine a coarse, fast model for selecting relevant sentences and a more expensive RNN for explaining/supporting answers and use this model to maintain or even improve the performance of state-of-the-art QA models simultaneously.
Lin et al. \cite{Lin:16} utilize a sentence-level selective attention to aggregate the information of all sentences to extract relational facts. They employ convolutional neural networks to embed the semantics of sentences.
Lin et al. \cite{lin:18} employs a paragraph selector to filter out those noisy paragraphs and a paragraph reader to extract the correct answer from those denoised paragraphs.
Shen et al. \cite{Shen:17} propose a distantly supervised open-domain question answering (DS-QA) system which uses information retrieval technique to obtain relevant text from Wikipedia as supporting facts for answers. 
Wang et al. \cite{wangsh:18} use reinforcement learning to train target paragraph selection and answer extraction jointly. They propose a re-ranking-based framework to make use of the evidence from multiple passages in open-domain QA, and perform evidence aggregation in existing open-domain QA datasets. 
Atanasova et al.~\cite{Generating_Fact_Checking} generate veracity explanations on available claim context, and show that veracity prediction
can be modeled jointly with veracity prediction and improves the performance of the veracity system. 
However, sentence-level evidences still have redundant noises. To guarantee continuous information, several whole sentences are selected out as supporting evidences instead of emphasizing clue words or phrases. 
Different from them, our research aims at token-level instead of sentence-level or paragraph-level to distill informative-yet-concise and human-readable supporting evidences. 
Human evaluation indicates that our distilled evidences are of high-quality which are informative to explain/support the input answers, concise without redundant noises and achieve better human  readability.

\textbf{Evidences extraction by neural networks for answer explanation}.
Besides, there are a series of research make answer-explanations by neural networks on complex datasets.
% They either apply KBs to make path searching, detecting entities and linking relations, or .
For example, Schuff et al.~\cite{schuff:20} propose a hierarchical model and a new regularization term to strengthen the answer-explanation coupling on multi-hop HOTPOTQA. 
Thayaparan et al.~\cite{Thayaparan:20} introduce a novel approach for answering and explaining multiple-choice science questions by reasoning on grounding and abstract inference chains. 
Wang et al.~\cite{HaiWang:19} extract evidence sentences for multiple-choice MRC based on inference and utilization of prior knowledge by distant supervision and deep probabilistic logic framework.
Jansen et al.~\cite{jansen:16} implement a fine-grained characterization of the knowledge and inference requirements on science exam QA. 
Tran et al.~\cite{tran:20} adopt an explainable, evidence-based memory network architecture, which learns to summarize the dataset and extract supporting evidences to make its decision on TrecQA and WikiQA. 
% However, some KBs-oriented explanations are composed of phrases corresponding to entities in structured KBs, which are hard to understand and support answers. And complicated relations among entities in knowledge graph also decrease the direct interpretability of this type of evidences. 
However, due to the end-to-end attribute of neural networks, the supporting evidences generated from neural networks are not traceable and user-friendly for explaining/supporting given answers. 
Different from their works, our solution GCED is built upon a pipelined principle based on three characteristics heuristically, improving the controllability and traceablility for the distilled evidences.
Moreover, Some research such as Atanasova et al.~\cite{A_Diagnostic_Study} aims to develop a comprehensive list of properties to evaluate exsting explainability techniques (simplification, gradient-based, etc).
Therefore, we could evaluate our GCED on this methods in the future.

\section{Conclusion and Future work}
The interpretability of question answering is a great challenge in NLP, which is expected to find supporting facts for given QA pairs so as to provide end-users a better understanding for why a specific prediction is regarded as the answer to that question. To automate the process of distilling high-quality evidences to explain/support given answers, we propose an Informative-yet-Concise Evidence Distillation algorithm (GCED) for answer explanation on QA. Empirical experiments elaborate the distilled evidences are informative to explain/support the input answers, concise without redundant noises and have human-like readability.
% Moreover, if our distilled evidences are based on ground-truth answers, they can substitute the contexts to improve the performance of QA models.
There is also some space to improve our algorithm. We plan to verify the proposed GCED on more datasets and models, improving its ability on understanding world knowledge/commonsense and speeding up the process of evidence distillation in the future.

\section{Acknowledgements}
% We thank the anonymous reviewers for their advice and suggestions for the manuscript. 
This work is supported by National Key Research and Development Project (No.2020AAA0109302), Shanghai Science and Technology Innovation Action Plan (No.19511120400) and Shanghai Municipal Science and Technology Major Project (No.2021SHZDZX0103).


\clearpage

 \begin{thebibliography}{00}
 \bibitem{Generating_Fact_Checking} Atanasova Pepa,  Simonsen Jakob Grue, Lioma Christina, and Augenstein Isabelle. 2020. Generating Fact Checking Explanations. In Proceedings of the 58th Annual Meeting of the Association for Computational Linguistics.7352--7364. 10.18653/v1/2020.acl-main.656.
 \bibitem{A_Diagnostic_Study} Atanasova Pepa,  Simonsen Jakob Grue, Lioma Christina, and Augenstein Isabelle. 2020. A Diagnostic Study of Explainability Techniques for Text Classification.
 In Proceedings of the 2020 Conference on Empirical Methods in Natural Language Processing (EMNLP).3256--3274.10.18653/v1/2020.emnlp-main.263.
 \bibitem{Big_Bird} Big Bird: Transformers for Longer Sequences.
 \bibitem{DEBERTA}DEBERTA: DECODING-ENHANCED BERT WITH DISENTANGLED ATTENTION.
 \bibitem{T5} Exploring the Limits of Transfer Learning with a Unified Text-to-Text Transformer.
  \bibitem{Knowledge_Guided_Text} Knowledge Guided Text Retrieval and Reading for Open Domain Question Answering.
 \bibitem{Longformer} Longformer: The Long-Document Transformer.
 \bibitem{The_probabilistic_relevance}The probabilistic relevance framework: Bm25 and beyond.
 \bibitem{RoBERTa} RoBERTa: A Robustly Optimized BERT Pretraining Approach.
 \bibitem{Berant:13} Jonathan Berant, Andrew Chou, Roy Frostig, and Percy Liang. 2013.Semantic Parsing on Freebase from Question-Answer Pairs. In Proceedings of the 2013
 Conference on Empirical Methods in Natural Language Processing. Association for Computational Linguistics, Seattle, Washington, USA, 1533–1544. https://www.aclweb.org/anthology/D13-1160
 \bibitem{Afrae:21} Ben Ahmed Mohamed Bghiel Afrae and Boudhir Anouar Abdelhakim. 2021.
Smart Sustainable Cities: A Chatbot Based on Question Answering System Passing by a Grammatical Correction for Serving Citizens.
 \bibitem{Avila:20}José Gilvan R. Maia Caio Viktor S. Avila, Wellington Franco and Vania M. P. Vidal.2020. CONQUEST: A Framework for Building Template-Based IQA Chatbots for Enterprise Knowledge Graphs.
 \bibitem{charniak:05} Eugene Charniak and Mark Johnson. 2005. Coarse-to-Fine n-Best Parsing and MaxEnt Discriminative Reranking. In Proceedings of the 43rd Annual Meeting of the
Association for Computational Linguistics (ACL’05). Association for Computational Linguistics, Ann Arbor, Michigan, 173–180. https://doi.org/10.3115/1219840.1219862
\bibitem{Chen:17} Danqi Chen, Adam Fisch, Jason Weston, and Antoine Bordes. 2017. Reading
Wikipedia to Answer Open-Domain Questions. In Proceedings of the 55th Annual Meeting of the Association for Computational Linguistics (Volume 1: Long Papers). Association for Computational Linguistics, Vancouver, Canada, 1870–1879. https://doi.org/10.18653/v1/P17-1171

\bibitem{Cheng:20}Hao Cheng, Ming-Wei Chang, Kenton Lee, and Kristina Toutanova. 2020.
Probabilistic Assumptions Matter: Improved Models for Distantly-Supervised Document-Level Question Answering. In Proceedings of the 58th Annual Meeting of the Association for Computational Linguistics. Association for Computational Linguistics, Online, 5657–5667. https://doi.org/10.18653/v1/2020.acl-main.501
\bibitem{Cheng:21} Hao Cheng, Xiaodong Liu, Lis Pereira, Yaoliang Yu, and Jianfeng Gao. 2021. Posterior Differential Regularization with f-divergence for Improving Model
 Robustness. In Proceedings of the 2021 Conference of the North American Chapter of the Association for Computational Linguistics: Human Language Technologies. Association for Computational Linguistics, Online, 1078–1089. https://www.aclweb.org/anthology/202 naacl-main.85
 \bibitem{choi:17} Eunsol Choi, Daniel Hewlett, Jakob Uszkoreit, Illia Polosukhin, Alexandre Lacoste, and Jonathan Berant. 2017. Coarse-to-Fine Question Answering for Long
Documents. In Proceedings of the 55th Annual Meeting of the Association for Computational Linguistics (Volume 1: Long Papers). Association for Computational Linguistics, Vancouver, Canada, 209–220. https://doi.org/10.18653/v1/P17-1020
\bibitem{chomsky:65} Noam Chomsky. 1965. Aspects of the Theory of Syntax. In MIT Press.
\bibitem{Kevin:20} Kevin Clark, Minh-Thang Luong, Quoc V. Le, and Christopher D. Manning. 2020. ELECTRA: Pre-training Text Encoders as Discriminators Rather Than Generators.
 In 8th International Conference on Learning Representations, ICLR 2020, Addis Ababa, Ethiopia, April 26-30, 2020. OpenReview.net. https://openreview.net/forum?id=r1xMH1BtvB
 \bibitem{collins:99} Michael Collins. 2003. Head-Driven Statistical Models for Natural Language Parsing. Computational Linguistics 29, 4 (2003), 589–637. https://doi.org/10.1162/
089120103322753356
 \bibitem{davies:02} Philip Davies. 2002. What is Evidence-based Education?. In British Journal of Educational Studies.
 \bibitem{devlin:18} Jacob Devlin, Ming-Wei Chang, Kenton Lee, and Kristina Toutanova. 2019. BERT: Pre-training of Deep Bidirectional Transformers for Language Understanding. In Proceedings of the 2019 Conference of the North American Chapter of the Association for Computational Linguistics: Human Language Technologies, Volume 1 (Long and Short Papers). Association for Computational Linguistics, Minneapolis, Minnesota,4171–4186. https://doi.org/10.18653/v1/N19-1423
 \bibitem{fellbaum:10} Christiane Fellbaum. 2010. WordNet. In Theory and Applications of Ontology: Computer Applications. 231–243.
 \bibitem{jansen:16} Peter Jansen, Niranjan Balasubramanian, Mihai Surdeanu, and Peter Clark.
 2016. What’s in an Explanation? Characterizing Knowledge and Inference
 Requirements for Elementary Science Exams. In Proceedings of COLING 2016,
 the 26th International Conference on Computational Linguistics: Technical Papers.
 The COLING 2016 Organizing Committee, Osaka, Japan, 2956–2965. https:
 //www.aclweb.org/anthology/C16-1278
 \bibitem{Jia:17}Robin Jia and Percy Liang. 2017. Adversarial Examples for Evaluating Reading
 Comprehension Systems. In Proceedings of the 2017 Conference on Empirical Methods in Natural Language Processing. Association for Computational Linguistics, Copenhagen, Denmark, 2021–2031. https://doi.org/10.18653/v1/D17-1215
 \bibitem{SpanBERT} Mandar Joshi, Danqi Chen, Yinhan Liu, Daniel S. Weld, Luke Zettlemoyer, and Omer Levy. 2020. SpanBERT: Improving Pre-training by Representing and Predicting Spans. Transactions of the Association for Computational Linguistics 8 (2020), 64–77. 
\bibitem{triviaqa:17} Mandar Joshi, Eunsol Choi, Daniel Weld, and Luke Zettlemoyer. 2017. TriviaQA: A large-scale Distantly Supervised Challenge Dataset for Reading Comprehension.
 In Proceedings of the 55th Annual Meeting of the Association for Computational Linguistics (Volume 1: Long Papers). Association for Computational Linguistics, Vancouver, Canada, 1601–1611. https://doi.org/10.18653/v1/P17-1147
 \bibitem{Kwiatkowski:13}Tom Kwiatkowski, Eunsol Choi, Yoav Artzi, and Luke Zettlemoyer. 2013. Scaling Semantic Parsers with On-the-Fly Ontology Matching. In Proceedings of the 2013
 Conference on Empirical Methods in Natural Language Processing. Association for Computational Linguistics, Seattle, Washington, USA, 1545–1556. https://www.aclweb.org/anthology/D13-1161
 \bibitem{Lan:20} Zhenzhong Lan, Mingda Chen, Sebastian Goodman, Kevin Gimpel, Piyush
 Sharma, and Radu Soricut. 2020. ALBERT: A Lite BERT for Self-supervised Learning of Language Representations. In 8th International Conference on Learning
 Representations, ICLR 2020, Addis Ababa, Ethiopia, April 26-30, 2020. OpenReview.net. https://openreview.net/forum?id=H1eA7AEtvS
 \bibitem{Patrick:20} Patrick S. H. Lewis, Ethan Perez, Aleksandra Piktus, Fabio Petroni, Vladimir Karpukhin, Naman Goyal, Heinrich Küttler, Mike Lewis, Wen-tau Yih, Tim
 Rocktäschel, Sebastian Riedel, and Douwe Kiela. 2020. Retrieval-Augmented Generation for Knowledge-Intensive NLP Tasks. In Advances in Neural Information Processing Systems 33: Annual Conference on Neural Information Processing Systems 2020, NeurIPS 2020, December 6-12, 2020, virtual, Hugo Larochelle, Marc’Aurelio Ranzato, Raia Hadsell, Maria-Florina Balcan, and Hsuan-Tien Lin (Eds.). https://proceedings.neurips.cc/paper/2020/hash/
 6b493230205f780e1bc26945df7481e5-Abstract.html
 \bibitem{lin:18} Yankai Lin, Haozhe Ji, Zhiyuan Liu, and Maosong Sun. 2018. Denoising Distantly Supervised Open-Domain Question Answering. In Proceedings of the 56th Annual
 Meeting of the Association for Computational Linguistics (Volume 1: Long Papers).
 Association for Computational Linguistics, Melbourne, Australia, 1736–1745.
 https://doi.org/10.18653/v1/P18-1161
 \bibitem{Lin:16} Yankai Lin, Shiqi Shen, Zhiyuan Liu, Huanbo Luan, and Maosong Sun. 2016.
 Neural Relation Extraction with Selective Attention over Instances. In Proceedings
 of the 54th Annual Meeting of the Association for Computational Linguistics (Volume
 1: Long Papers). Association for Computational Linguistics, Berlin, Germany,
 2124–2133. https://doi.org/10.18653/v1/P16-1200
 \bibitem{sackett:97} David L.Sackett. 1997. Evidence-based medicine. In Seminars in Perinatology.3–5.
 \bibitem{Min:19}Sewon Min, Danqi Chen, Hannaneh Hajishirzi, and Luke Zettlemoyer. 2019. A
 Discrete Hard EM Approach for Weakly Supervised Question Answering. In Proceedings of the 2019 Conference on Empirical Methods in Natural Language Processing and the 9t International Joint Conference on Natural Language Processing (EMNLP-IJCNLP). Association for Computational Linguistics, Hong Kong, China,2851–2864. https://doi.org/10.18653/v1/D19-1284
 \bibitem{min:18} Sewon Min, Victor Zhong, Richard Socher, and Caiming Xiong. 2018. Efficient
 and Robust Question Answering from Minimal Context over Documents. In
 Proceedings of the 56th Annual Meeting of the Association for Computational
 Linguistics (Volume 1: Long Papers). Association for Computational Linguistics,
 Melbourne, Australia, 1725–1735. https://doi.org/10.18653/v1/P18-1160
 \bibitem{Jeong:20} Su-Jeong Song Min-Su Jeong and Yun-Gyung Cheong. 2020. Developing a Korean QA Chatbot for a Medical Domain Using a Multi-step Retrieval Approach.
 \bibitem{Mitra:20} Rajarshee Mitra, Manish Gupta, and Sandipan Dandapat. 2020. Transformer
 Models for Recommending Related Questions in Web Search. In CIKM ’20: The
 29th ACM International Conference on Information and Knowledge Management,
 Virtual Event, Ireland, October 19-23, 2020, Mathieu d’Aquin, Stefan Dietze, Claudia
 Hauff, Edward Curry, and Philippe Cudré-Mauroux (Eds.). ACM, 2153–2156.
 https://doi.org/10.1145/3340531.3412067
 \bibitem{Thayaparan:20} Marco Valentino Mokanarangan Thayaparan and André Freitas. 2020. ExplanationLP: Abductive Reasoning for Explainable Science Question Answering.
 \bibitem{harden:09} Graham Buckley R. M. Harden, Janet Grant and I. R. Hart. 2009. BEME Guide No.1: Best Evidence Medical Education. In Medical Teacher.
 \bibitem{Rajani:19} Nazneen Fatema Rajani, Bryan McCann, Caiming Xiong, and Richard Socher. 2019. Explain Yourself! Leveraging Language Models for Commonsense Reasoning.
 In Proceedings of the 57th Annual Meeting of the Association for Computational
 Linguistics. Association for Computational Linguistics, Florence, Italy, 4932–4942.
 https://doi.org/10.18653/v1/P19-1487
 \bibitem{squad:18} Pranav Rajpurkar, Robin Jia, and Percy Liang. 2018. Know What You Don’t
 Know: Unanswerable Questions for SQuAD. In Proceedings of the 56th Annual
 Meeting of the Association for Computational Linguistics (Volume 2: Short Papers).
 Association for Computational Linguistics, Melbourne, Australia, 784–789. https:
 //doi.org/10.18653/v1/P18-2124
 \bibitem{squad:16} Pranav Rajpurkar, Jian Zhang, Konstantin Lopyrev, and Percy Liang. 2016.
 SQuAD: 100,000+ Questions for Machine Comprehension of Text. In Proceedings of the 2016 Conference on Empirical Methods in Natural Language Processing. Association for Computational Linguistics, Austin, Texas, 2383–2392.https://doi.org/10.18653/v1/D16-1264
 \bibitem{schuff:20} Hendrik Schuff, Heike Adel, and Ngoc Thang Vu. 2020. F1 is Not Enough! Models and Evaluation Towards User-Centered Explainable Question Answering. In
 Proceedings of the 2020 Conference on Empirical Methods in Natural Language
 Processing (EMNLP). Association for Computational Linguistics, Online, 7076–
 7095. https://doi.org/10.18653/v1/2020.emnlp-main.575
 \bibitem{Shen:17} Yelong Shen, Po-Sen Huang, Jianfeng Gao, and Weizhu Chen. 2017. ReasoNet:
 Learning to Stop Reading in Machine Comprehension. In Proceedings of the
 23rd ACM SIGKDD International Conference on Knowledge Discovery and Data
 Mining, Halifax, NS, Canada, August 13 - 17, 2017. ACM, 1047–1055. https://doi.org/10.1145/3097983.3098177
 \bibitem{Shou:20} Linjun Shou, Shining Bo, Feixiang Cheng, Ming Gong, Jian Pei, and Daxin Jiang. 2020. Mining Implicit Relevance Feedback from User Behavior for Web Question
 Answering. In KDD ’20: The 26th ACM SIGKDD Conference on Knowledge Discovery
 and Data Mining, Virtual Event, CA, USA, August 23-27, 2020, Rajesh Gupta,
 Yan Liu, Jiliang Tang, and B. Aditya Prakash (Eds.). ACM, 2931–2941. https://dl.acm.org/doi/10.1145/3394486.3403343
 \bibitem{tran:20} Quan Hung Tran, Nhan Dam, Tuan Lai, Franck Dernoncourt, Trung Le, Nham
 Le, and Dinh Phung. 2020. Explain by Evidence: An Explainable Memorybased Neural Network for Question Answering. In Proceedings of the 28th International Conference on Computational Linguistics. International Committee on Computational Linguistics, Barcelona, Spain (Online), 5205–5210. https://doi.org/10.18653/v1/2020.coling-main.456
 \bibitem{HaiWang:19} Hai Wang, Dian Yu, Kai Sun, Jianshu Chen, Dong Yu, David McAllester, and Dan Roth. 2019. Evidence Sentence Extraction for Machine Reading Comprehension.
 In Proceedings of the 23rd Conference on Computational Natural Language Learning
 (CoNLL). Association for Computational Linguistics, Hong Kong, China, 696–707.
 https://doi.org/10.18653/v1/K19-1065
 \bibitem{wangsh:18} Shuohang Wang, Mo Yu, Jing Jiang, Wei Zhang, Xiaoxiao Guo, Shiyu Chang,
 Zhiguo Wang, Tim Klinger, Gerald Tesauro, and Murray Campbell. 2018. Evidence
 Aggregation for Answer Re-Ranking in Open-Domain Question Answering. In
 6th International Conference on Learning Representations, ICLR 2018, Vancouver,
 BC, Canada, April 30 - May 3, 2018, Conference Track Proceedings. OpenReview.net.
 https://openreview.net/forum?id=rJl3yM-Ab
 \bibitem{weisz:10} John R. Weisz and Alan E. Kazdin. 2010. Evidence-Based Psychotherapies for Children and Adolescents. New York.
 \bibitem{Yamada:20} Ikuya Yamada, Akari Asai, Hiroyuki Shindo, Hideaki Takeda, and Yuji Matsumoto.2020. LUKE: Deep Contextualized Entity Representations with Entity-aware Selfattention. In Proceedings of the 2020 Conference on Empirical Methods in Natural
 Language Processing (EMNLP). Association for Computational Linguistics, Online,
 6442–6454. https://doi.org/10.18653/v1/2020.emnlp-main.523
 \bibitem{Yangw:19} Wei Yang, Yuqing Xie, Aileen Lin, Xingyu Li, Luchen Tan, Kun Xiong, Ming
 Li, and Jimmy Lin. 2019. End-to-End Open-Domain Question Answering with BERTserini. In Proceedings of the 2019 Conference of the North American Chapter of the Association for Computational Linguistics (Demonstrations). Association for Computational Linguistics, Minneapolis, Minnesota, 72–77. https://doi.org/10.18653/v1/N19-4013
 \bibitem{yi:15} Yi Yang, Wen-tau Yih, and Christopher Meek. 2015. WikiQA: A Challenge Dataset for Open-Domain Question Answering. In Proceedings of the 2015 Conference on
 Empirical Methods in Natural Language Processing. Association for Computational
 Linguistics, Lisbon, Portugal, 2013–2018. https://doi.org/10.18653/v1/D15-1237
 \bibitem{Yang:19} Zhilin Yang, Zihang Dai, Yiming Yang, Jaime G. Carbonell, Ruslan Salakhutdinov, and Quoc V. Le. 2019. XLNet: Generalized Autoregressive Pretraining for
 Language Understanding. In Advances in Neural Information Processing Systems
 32: Annual Conference on Neural Information Processing Systems 2019, NeurIPS
 2019, December 8-14, 2019, Vancouver, BC, Canada, Hanna M. Wallach, Hugo
 Larochelle, Alina Beygelzimer, Florence d’Alché-Buc, Emily B. Fox, and Roman
 Garnett (Eds.). 5754–5764.
 \bibitem{Yih:15} Wen-tau Yih, Ming-Wei Chang, Xiaodong He, and Jianfeng Gao. 2015. Semantic
 Parsing via Staged Query Graph Generation: Question Answering with Knowledge Base. In Proceedings of the 53rd Annual Meeting of the Association for Computational Linguistics and the 7th International Joint Conference on Natural Language Processing (Volume 1: Long Papers). Association for Computational Linguistics, Beijing, China, 1321–1331. https://doi.org/10.3115/v1/P15-1128
 \bibitem{Yin:16} Wenpeng Yin, Hinrich Schütze, Bing Xiang, and Bowen Zhou. 2016. ABCNN:
 Attention-Based Convolutional Neural Network for Modeling Sentence Pairs. Transactions of the Association for Computational Linguistics 4 (2016), 259–272.
 
\end{thebibliography}
\end{document}